\documentclass[10pt,twocolumn,letterpaper]{article}

\usepackage[pagenumbers]{iccv} 
\usepackage{multirow}
\usepackage{hyperref}
\usepackage{tikz}
\usetikzlibrary{shapes, arrows.meta, positioning, fit, calc}
\usepackage{adjustbox}

\definecolor{iccvblue}{rgb}{0.21,0.49,0.74}

\def\paperID{*****} 
\def\confName{ICCV}
\def\confYear{2025}

\title{Deep Learning for Crack Detection: A Review of Learning Paradigms, Generalizability, and Datasets}

\author{
Xinan~Zhang$^{1}$ \quad Haolin~Wang$^{1}$ \quad Yung-An~Hsieh$^{1}$ \quad Zhongyu~Yang$^{1}$ \quad Anthony~Yezzi$^{1}$ \quad Yi-Chang~Tsai$^{1}$\\
Georgia Institute of Technology$^{1}$
\thanks{This work has been submitted to the IEEE for possible publication. Copyright may be transferred without notice, after which this version may no longer be accessible.}
}
\addtocounter{footnote}{-1}

\begin{document}
\maketitle
\begin{abstract}
Crack detection plays a crucial role in civil infrastructures, including inspection of pavements, buildings, etc., and deep learning has significantly advanced this field in recent years. While numerous technical and review papers exist in this domain, emerging trends are reshaping the landscape. These shifts include transitions in learning paradigms (from fully supervised learning to semi-supervised, weakly-supervised, unsupervised, few-shot, domain adaptation and fine-tuning foundation models), improvements in generalizability (from single-dataset performance to cross-dataset evaluation), and diversification in dataset acquisition (from RGB images to specialized sensor-based data). In this review, we systematically analyze these trends and highlight representative works. Additionally, we introduce a new annotated dataset collected with 3D laser scans, 3DCrack, to support future research and conduct extensive benchmarking experiments to establish baselines for commonly used deep learning methodologies, including recent foundation models. Our findings provide insights into the evolving methodologies and future directions in deep learning-based crack detection. Project page: \href{https://github.com/nantonzhang/Awesome-Crack-Detection}{https://github.com/nantonzhang/Awesome-Crack-Detection}.
\end{abstract}    
\section{Introduction}

\begin{figure}[htbp]
    \centering
    \begin{subfigure}[t]{0.48\linewidth}
        \centering
        \includegraphics[width=\linewidth]{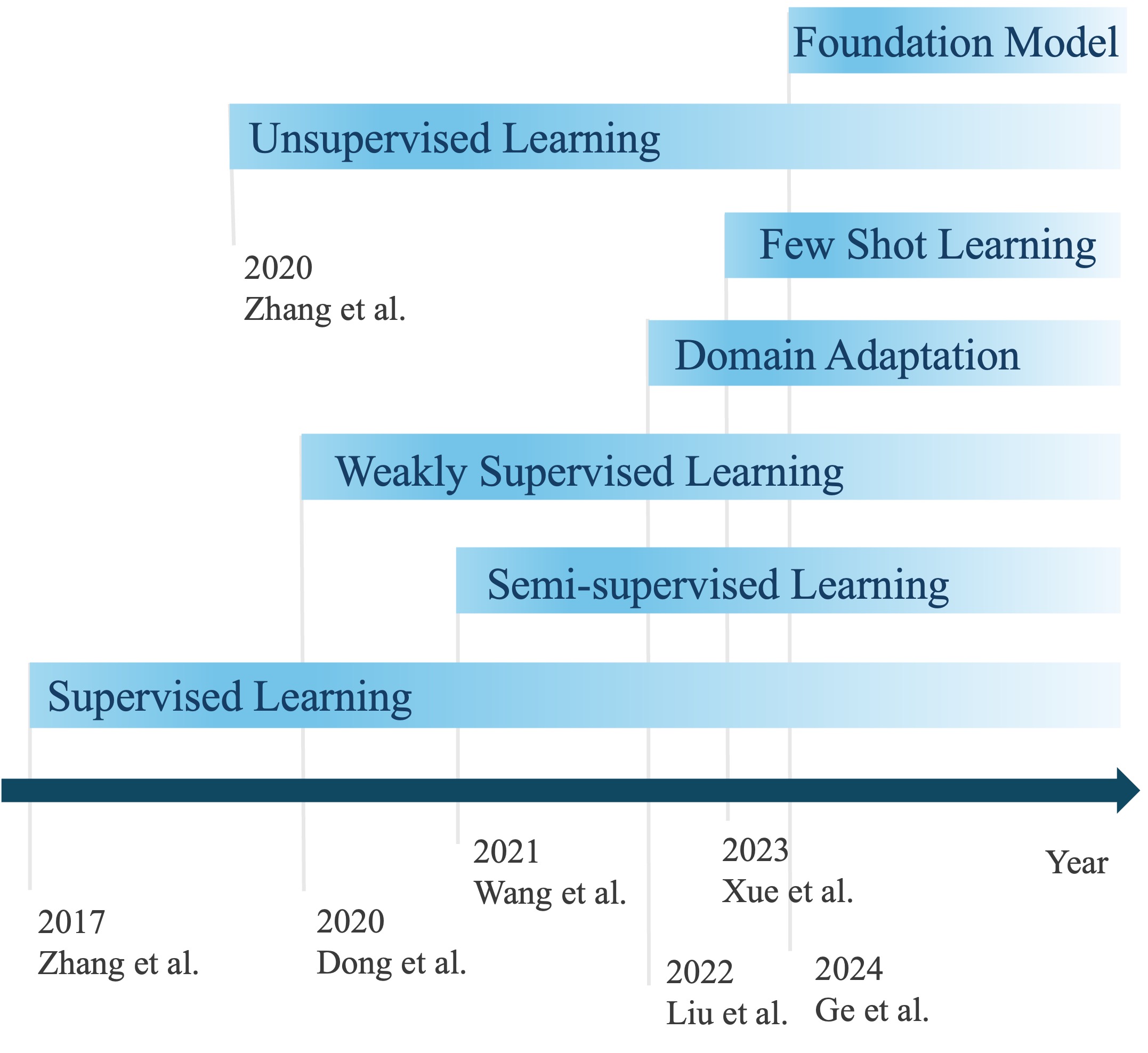}
        \caption{}
        \label{fig:teaser1}
    \end{subfigure}
    \hfill
    \begin{subfigure}[t]{0.48\linewidth}
        \centering
        \includegraphics[width=\linewidth]{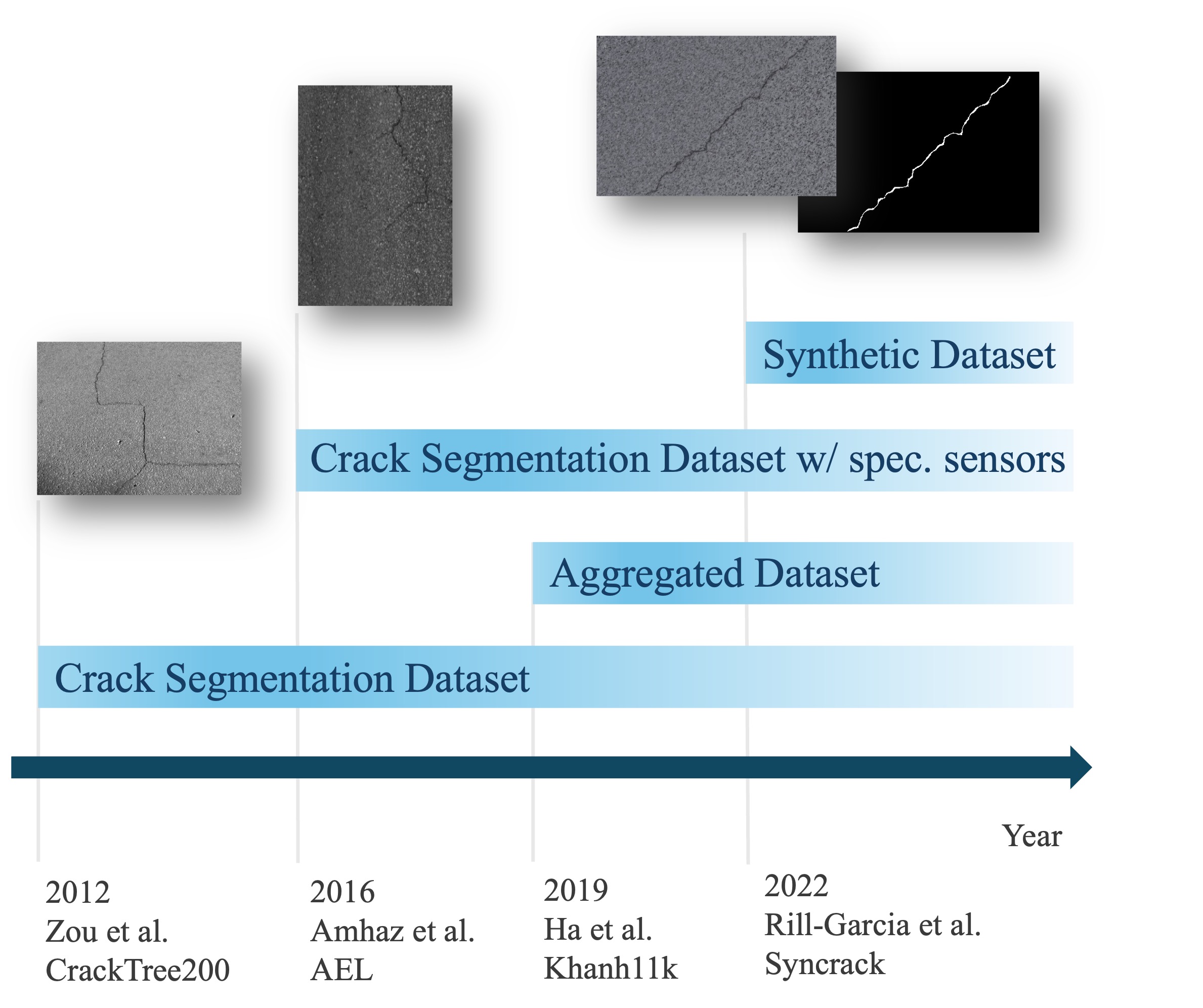}
        \caption{}
        \label{fig:teaser2}
    \end{subfigure}
    \caption{Timelines of (a) emerging learning paradigms in deep learning-based crack segmentation and (b) releases of publicly available datasets for crack segmentation based on different categories. Both (a) and (b) highlight first representative works.}
    \label{fig:combined}
\end{figure}

\begin{figure*}[ht]
\centering
\begin{adjustbox}{width=\linewidth}
\begin{tikzpicture}[
    level 1/.style={sibling distance=40mm},
    level 2/.style={sibling distance=40mm},
    every node/.style={font=\small},
    edge from parent/.style={draw, -latex},
    concept/.style={
      rectangle, rounded corners,
      draw=black, fill=blue!10,
      text width=3cm, align=center, font=\small
  },
  leaf/.style={
      rectangle, rounded corners,         
      draw=black, fill=gray!08,
      text width=3.5cm, align=left,    
      minimum height=1.6cm,              
      font=\footnotesize,
      inner sep=3pt
  },
]

\node [concept] {Deep Learning for Crack Detection}
    child [level distance=20mm] {node [concept] {Background: Previous Reviews (Sec.~\ref{sec:background})}
        child {node [leaf] {1) IP/ML/DL Based Review\\ 2) Task Based Review\\ 3) Pipeline Based Review}}
    }
    child [level distance=20mm] {node [concept] {Learning Paradigms \& Generalizability \\\small(Sec.~\ref{sec:learning})}
        child {node [leaf] {1) Seven Main Paradigms, with Details in Fig.~\ref{fig:taxonomy}\\ 2) Generalizability Analysis}}
    }
    child [level distance=20mm] {node [concept] {Datasets (Sec.~\ref{sec:datasets})}
        child {node [leaf] {1) Review of Existing Datasets \& Analysis\\ 2) Ours: 3DCrack}}
    }
    child [level distance=20mm] {node [concept] {Experiments \& Findings (Sec.~\ref{sec:experiments})}
        child {node [leaf] {1) Supervised \& Semi-Supervised Experiments\\ 2) Generalizable Experiments}}
    }
    child [level distance=20mm] {node [concept] {Open Challenges\\\small (Sec.~\ref{sec:open})}
        child {node [leaf] {Challenge \& Opportunity in\\ 1) Learning Paradigms\\ 2) Generalizability \& Model Development\\ 3) Dataset}}
    }
    child [level distance=20mm] {node [concept] {Conclusion (Sec.~\ref{sec:conclusion})}
        child {node [leaf] {1) Shifts in Learning Paradigms\\ 2) Emergence of Generalizability\\ 3) Diversification in Datasets}}
    };

\end{tikzpicture}
\end{adjustbox}
\caption{Structure of this review paper.}
\label{fig:structure}
\end{figure*}
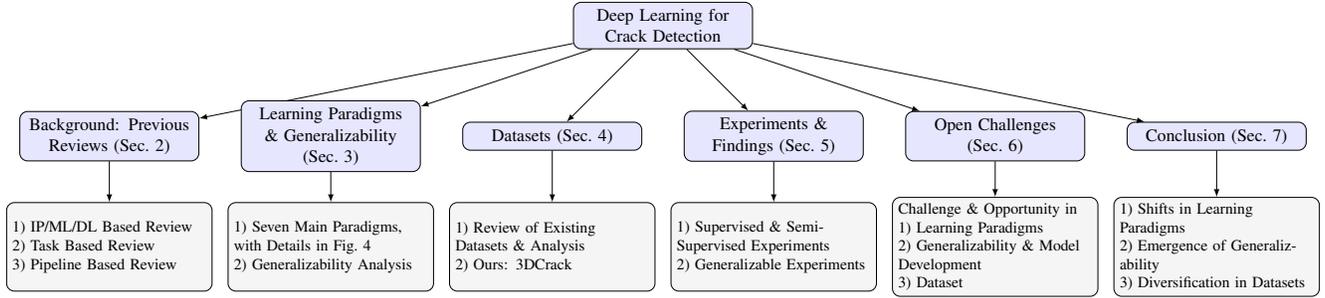

Crack detection is a crucial task in civil infrastructures ~\cite{zhang2017automated,zhang2018deep,hsieh2020machine}, including pavements, buildings, etc. Timely and accurate identification of cracks allows for proactive maintenance, reducing repair costs and preventing severe structural failures. As a result, crack detection has been a significant area of research, evolving through various stages over the past few decades~\cite{mohan2018crack,hsieh2020machine}.

Initially, crack detection relied on manual surveys, where trained inspectors visually assessed pavement conditions. While effective, this approach was labor-intensive, time-consuming, and prone to subjective error. To improve efficiency, researchers developed methods based on traditional image processing~\cite{mohan2018crack,hsieh2020machine,cao2020review}, leveraging techniques such as edge detection, intensity thresholding, and morphological operations to extract crack features from images. But these rule-based methods often struggled with variations in lighting, pavement texture, and crack patterns. 

The introduction of machine learning (ML) marked a major advancement, as ML models could learn feature representations from labeled datasets, improving crack detection accuracy. Yet, traditional ML methods still required pre-extracted initial features and lacked the scalability needed for real-world deployment~\cite{mohan2018crack,hsieh2020machine,cao2020review}. With the rise of deep learning (DL), represented by the wide adoption of convolutional neural networks (CNN) and transformers, crack detection entered a new era, enabling automated feature extraction and significantly enhancing detection performance~\cite{zhang2017automated,zhang2018deep}. 

DL has been applied to crack detection through various approaches, including classification, object detection, semantic segmentation, etc.~\cite{munawar2021image,hsieh2020machine,cao2020review,kheradmandi2022critical} Classification models determine whether an image contains cracks, while object detection methods localize cracks by drawing bounding boxes around them. However, both approaches lack the fine-grained precision needed for accurate crack characterization. Among these methods, semantic segmentation stands out as it enables precise pixel-wise identification of cracks, making it ideal for detailed crack identification and quantification~\cite{zhang2017automated,zhang2018deep,hsieh2020machine}. Given its importance and growing adoption in the field, this review focuses specifically on deep learning-based semantic segmentation methods for crack detection. As presented in Fig.~\ref{fig:combined}, the field has witnessed rapid progress in recent years, with numerous studies exploring different learning paradigms and datasets, from supervised learning to unsupervised learning; from pursuit of performance on single datasets to generalizability; from RGB camera data to data by specialized sensors, etc. 

\begin{figure}
    \centering
    \includegraphics[width=1\linewidth]{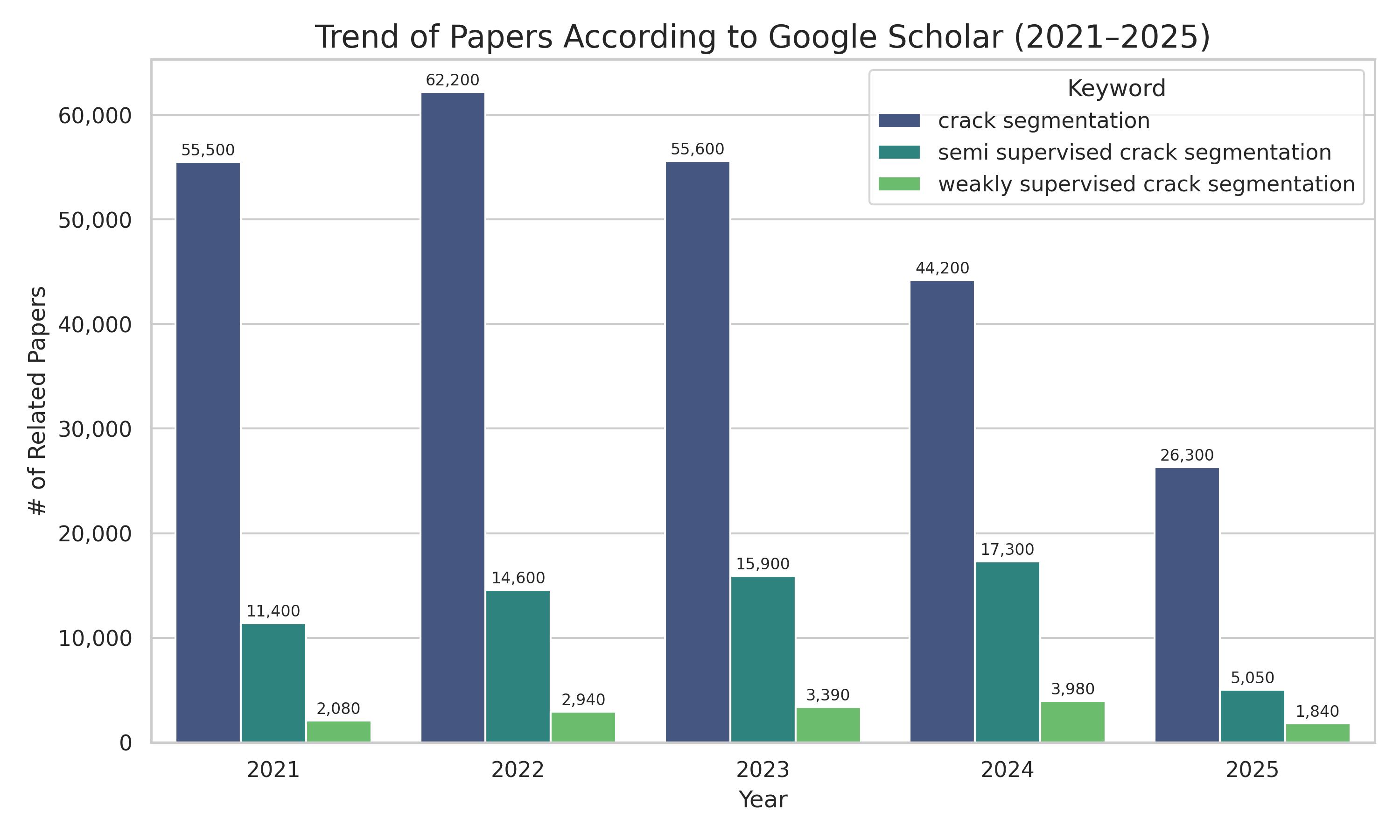}
    \caption{Trend of related papers in the recent five years by Google Scholar with different keywords, as of June 3rd, 2025. Although the total number of papers related to crack segmentation declines after 2022, non-fully supervised learning paradigms such as semi-supervised and weakly-supervised learning have gained increasing attention in this domain.}
    \label{fig:trend}
\end{figure}

While several review papers already exist as presented in Tab.~\ref{tab:reviewpapers}, a systematic analysis of the latest trends is still in urgent need. Notably, three key trends in this field should be captured to make a thorough review:

\begin{enumerate}
    \item \textbf{Shifts in Learning Paradigms.} As shown in Fig.~\ref{fig:trend}, the focus of this domain has begun shifting toward data-efficient learning paradigms such as semi-supervised learning, which have introduced new challenges and opportunities. These shifts are driven by the growing need to overcome the challenges such as limited annotated data, domain gap, etc. Analyzing such developments in different learning paradigms, including supervised learning (SL), semi-supervised learning (SSL), weakly-supervised learning (WSL), unsupervised learning (USL), few-shot learning (FSL), domain adaptation (DA) and parameter-efficient fine-tuning (PEFT) foundation models, could provide clear and unique guidance for feature research and implementation. These shifts mirror trends in the broader CV community, while incorporating designs tailored to crack-specific features.
    \item \textbf{Emergence of Generalizability.} Crack detection models have traditionally been trained and evaluated on single datasets, limiting their real-world zero-shot applicability. Alongside the trends in learning paradigms and dataset development, recent efforts have been made to enhance generalizability across multiple datasets, which should be included in the systematic review for insights into frontiers in the field.
    \item \textbf{Diversification in Datasets.} Early works conducted crack segmentation with standard RGB images, but the field has started adopting professional sensor-based datasets (e.g., 3D laser scans), which offer more reliable and robust imaging for crack detection under diverse conditions. Availability of aggregated datasets and synthetic datasets also creates new opportunities for DL-based crack segmentation. Analysis of characteristics of different datasets is beneficial for bridging the gap between research practice and real-world implementation.
\end{enumerate}

Subsequently, in this review, we analyze these trends by examining recent representative works, and providing a category-specific, structured analysis of learning paradigms and their generalizability. The currently available datasets are also systematically reviewed. Additionally, we introduce a new high-resolution, large-scale, industry-grade dataset for crack detection, aimed at bridging the gap between research and real-world implementation. We also benchmark widely used deep learning models, including recent foundation models, across different learning settings with unique findings. Our goal is to provide a comprehensive and up-to-date perspective on DL-based crack detection, guiding future research in this rapidly evolving domain. This paper is structured as shown in Fig.~\ref{fig:structure}. To conclude, our contributions of this review paper are as follows:
\begin{enumerate}
    \item Three aforementioned trends are comprehensively studied with review on recent representative works and category-specific structured analysis for learning paradigms, discussing the strengths and generalizability for each category, in the domain of crack detection.
    \item To facilitate this systematic study and future research, we introduce a new high-resolution, large-scale, industry-grade dataset \textit{3DCrack} for crack semantic segmentation, which has higher resolution, more diverse pavement conditions compared with similar datasets.
    \item We summarize the current challenges and unique opportunities for future research and implementation, with benchmarking of the representative baseline models.
\end{enumerate}

\section{Background}
\label{sec:background}
As shown in Tab.~\ref{tab:reviewpapers}, early review papers in the field of crack detection primarily focused on image-processing (IP) techniques~\cite{mohan2018crack,wang2019pavement}. These approaches, favored for their simplicity and interpretability, typically relied on handcrafted features such as edge detection, thresholding, and texture analysis. As computational resources and annotated data became more available, the field witnessed a shift toward ML and DL approaches. This evolution is reflected in the review literature, which has transitioned from IP-centric surveys to more recent studies centered on data-driven and learning-based methods~\cite{wang2019pavement,cao2020review,hsieh2020machine,munawar2021image}. This progression highlights both technological advances and an increasing demand for scalable, accurate, and automated crack detection systems.

Recent reviews have adopted a range of perspectives. Some categorize methods based on their underlying approaches—such as image classification (determining the presence of cracks), object detection (localizing cracks using bounding boxes), and semantic segmentation (assigning pixel-wise labels for crack regions)~\cite{hamishebahar2022comprehensive,li2022review,yuan2024review}. Others take a pipeline-oriented view, covering the entire workflow from data acquisition (e.g., drones, smartphones, or laser sensors) to crack identification, quantification, and decision support~\cite{kheradmandi2022critical,ai2023computer,yuan2024review}. A third line of reviews emphasizes computational aspects, including network architecture design, feature representation, and model optimization strategies~\cite{li2022review,nguyen2023deep}.

\begin{figure}
    \centering
    \includegraphics[width=0.9\linewidth]{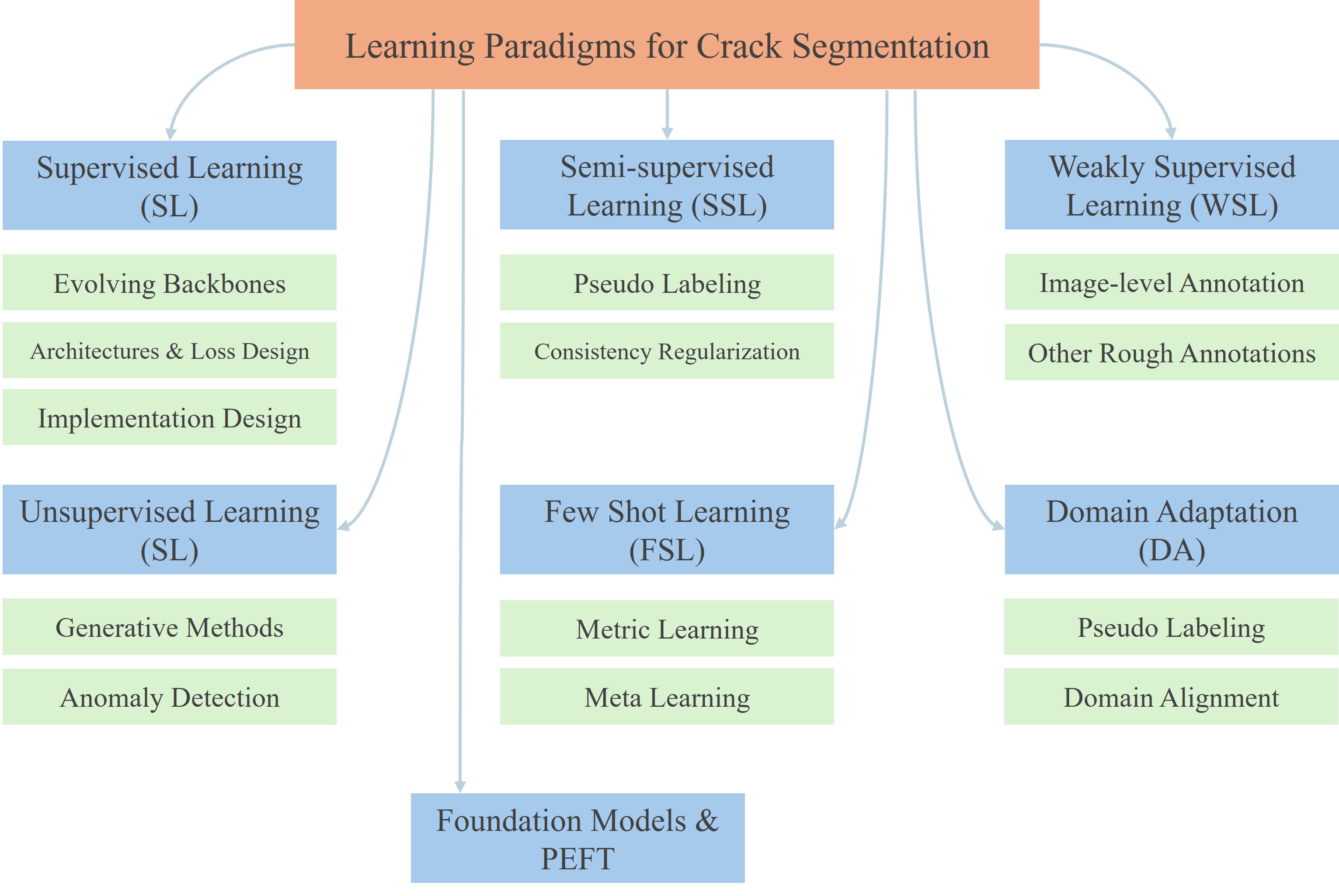}
    \caption{Review based on taxonomy of learning paradigms in Sec.~\ref{sec:learning}. Green boxes represent subgroups under each learning paradigm in each blue box.}
    \label{fig:taxonomy}
\end{figure}

However, no existing review has comprehensively analyzed the three key trends in a hierarchical structure—\textbf{Shifts in Learning Paradigms}, \textbf{Emergence of Generalizability}, and \textbf{Diversification in Datasets}. This paper aims to fill this gap by providing an in-depth review as shown in Fig.~\ref{fig:structure} \&~\ref{fig:taxonomy}, and discussion on these three focal areas, which are fundamental to the advancement of DL-based crack detection.

\begin{table}
    \centering
    \caption{Summary of review papers related to crack segmentation. SL, SSL, WSL, DA, FSL, USL, and FM stand for Supervised Learning, Semi-Supervised Learning, Weakly-Supervised Learning, Domain Adaptation, Few-Shot Learning, Unsupervised Learning, and Foundation Models, respectively.}
    \label{tab:reviewpapers}
    \large
    \renewcommand\arraystretch{1.5}
    \resizebox{\columnwidth}{!}{%
    \begin{tabular}{cp{10cm}p{3cm}}
        \toprule
         Review Paper&  Description& Learning Paradigms\\
        \midrule
         Mohan et al. (2018)~\cite{mohan2018crack}&  Literature presents different techniques to automatically identify the crack and its depth using image processing techniques.& None (IP Only)\\
         
         Wang et al. (2019)~\cite{wang2019pavement}&  A number of literature in this research topic are collected  for summarizing the research artwork status, and giving a review of the pavement crack image acquisition methods and 2D crack extraction algorithms. & SL\\
         
         Cao et al. (2020)~\cite{cao2020review}&  This paper reviews the three major types of methods used in road cracks detection: image processing, machine learning, and 3D imaging-based methods. & SL\\      
         
         Hsieh et al. (2020)~\cite{hsieh2020machine} & Authors organize and provide up-to-date information on ML-based crack detection algorithms for researchers to more efficiently seek potential focus and direction. & SL\\
         
         Munawar et al. (2021) ~\cite{munawar2021image}& This paper provides a review of image-based crack detection techniques that implement image processing and/or machine learning.&SL\\
         
         Ali et al. (2022) ~\cite{ali2022structural}& A review of CNN implementation on civil structure crack detection, highlighting significant research for crack detection through classification and segmentation. &SL\\

         Hamishebahar et al. (2022) ~\cite{hamishebahar2022comprehensive}& A comprehensive literature review of deep learning-based crack detection studies and the contributions they have made to the field is presented. &SL\\
         
         Li et al. (2022) ~\cite{li2022review}& It presents a comprehensive thematic survey of DL-based CIS techniques. Our review offers several contributions to the CIS area. &SL, WSL, USL\\
         
         Kheradmandi et al.  (2022)~\cite{kheradmandi2022critical}& This literature review focuses heavily on three major types of approaches in the field of image segmentation, namely thresholding-based, edge-based, and data-driven-based methods. &SL\\

         Zhou et al. (2023)~\cite{zhou2023deep}& It reviews recent developments in deep learning-based methods for crack segmentation and investigates the impact from different image types. & SL\\
         
         Nguyen et al. (2023)~\cite{nguyen2023deep}& This study has identified the bigger picture of DL methods for crack identification in asphalt pavement. The authors evaluated several DL-based crack identification algorithms from the literature. & SL, USL\\
         
         Ai et al. (2023)~\cite{ai2023computer}& This paper provides a comprehensive review of the research progress and prospects in computer vision frameworks for crack detection of civil infrastructures from multiple materials, including asphalt, concrete, and metal-like materials. &SL\\
         
         Yuan et al. (2024)~\cite{yuan2024review}& Based on the main research methods of the 120 documents, we classify them into three crack detection methods: fusion of traditional methods and deep learning, multimodal data fusion, and semantic image understanding. &SL\\
         
         Xu et al. (2024)~\cite{yang2024few}& This article systematically summarizes recent advances in FSL algorithms and the corresponding applications in SHD for civil infrastructure. &FSL\\
         
         Amirkhani et al. (2024)~\cite{amirkhani2024visual}& This survey aims to study the recent progress of vision-based concrete bridge defect classification and detection in the deep learning era.&SL, SSL, WSL, USL\\
         
         \textbf{Ours (2025)}& DL for crack segmentation is the focus of this survey, with trends in learning paradigm, generalizability, and dataset comprehensively reviewed and categorized. A professional dataset, 3DCrack, is released. &SL, SSL, WSL, DA, FSL, USL, FM\\
 \bottomrule
    \end{tabular}
    }
    
\end{table}

\section{Learning Paradigms \& Generalizability}
\label{sec:learning}

\begin{figure*}
    \centering
    \includegraphics[width=1\linewidth]{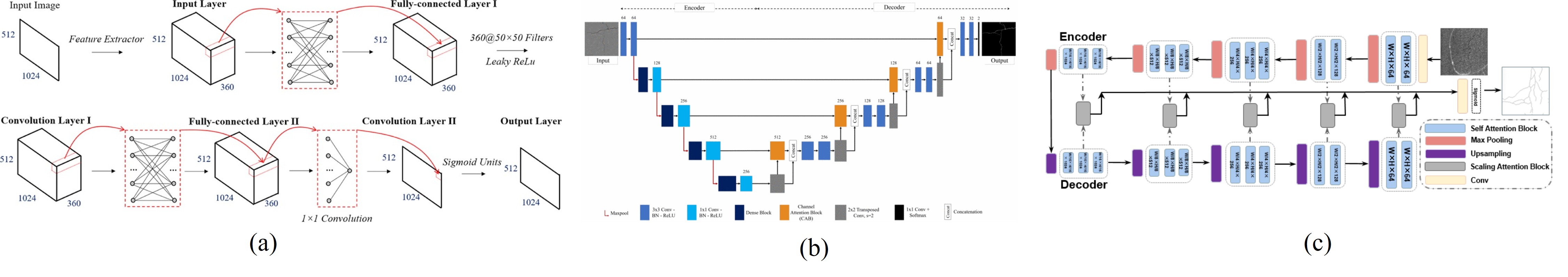}
    \caption{Evolution of supervised learning in crack segmentation regarding backbones and architectures. From (a) CrackNet~\cite{zhang2017automated} using the most fundamental CNN, to (b) DAUNet~\cite{hsieh2021dau} utilizing an encoder-decoder architecture, (c) CrackFormer~\cite{liu2021crackformer} applying transformers, evolution has been witnessed in the deeper backbone and more intricate architecture.}
    \label{fig:sl}
\end{figure*}

Crack detection by segmentation is fundamentally a task of semantic segmentation in the field of computer vision. The primary objective is to divide the input image into regions with distinct semantic meanings—in this case, differentiating between crack and non-crack areas at the pixel level. Unlike traditional classification or object detection, semantic segmentation provides fine-grained, pixel-wise predictions, making it particularly suitable for infrastructure inspection tasks such as crack detection, where precise detection of geometry is critical.

With the advent of deep learning, this area has witnessed rapid progress. CNN and, more recently, transformer-based models have been extensively adopted to capture complex spatial patterns and structural variations of cracks across diverse environments. These deep learning-based methods have significantly outperformed classical image processing techniques in terms of both accuracy and robustness~\cite{hsieh2020machine,cao2020review,li2022review}.

Alongside this progress, a variety of learning paradigms have emerged, each addressing different practical challenges, including improving performance, boosting generalization, handling limited labeled data, etc. These paradigms consist of supervised learning, in which models are trained on pixel-level fully annotated datasets; weakly-supervised approaches, which reduce annotation burden by using coarser labels such as scribbles or image-level labels; semi-supervised methods that leverage both labeled and unlabeled data; domain adaptation which aims to adapt models across different domains where training and test datasets fall into different data distributions; few-shot learning, which enables generalization to new tasks with only a few labeled examples; and unsupervised learning, which discovers latent structures in fully unlabeled data. Additionally, motivated by recent advances in generalizability in the broader CV community, such as Contrastive Language-Image Pre-training (CLIP)~\cite{radford2021learning} and Segment Anything Model (SAM)~\cite{kirillov2023segment}, generalizable methods based on foundation models and parameter-efficient fine-tuning (PEFT) techniques~\cite{xu2023parameter} designed for fine-tuning large models have started to emerge in crack segmentation.

Given the increasing popularity and diversity of these approaches, this section provides a structured review of the major learning paradigms in DL-based crack segmentation. It also discusses the advantages of each paradigm, particularly with regard to their potential for generalization.

\subsection{Supervised Learning}
In fully supervised crack segmentation, the problem can be formulated as follows. Given a fully pixel-level annotated dataset $\mathcal{D}_L = \{(x_1, y_1), \ldots, (x_l, y_l)\}$, the objective is $\min_\theta \frac{1}{n} \sum_{i=1}^{n} \mathcal{L}(f_\theta(x_i), y_i)$, where $f_\theta$ is the learned segmentation network which aims to segment the cracking area. 

This domain has witnessed notable progress in recent years. The advancement can be categorized into three key stages, including 1) stronger backbones, 2) more intricate structure setup and loss design for better performance, and 3) specific designs for field implementation.

\textbf{Evolving Backbones.} In DL-based crack segmentation, the transition of backbone selection in models has been witnessed from CNN to transformer. Initially, shallow CNNs, often with fewer layers and simpler structures, are employed as backbones ~\cite{zhang2017automated,zhang2018deep,liu2019deepcrack}. These early CNN-based models already demonstrated superior performance compared to traditional ML and image processing techniques, primarily due to their strong ability to learn feature representations directly from raw images or pre-processed preliminary features as shown in Fig.~\ref{fig:sl}(a). As this field progressed and more diverse datasets became available, as shown in Tabs.~\ref{tab:dataset_1} \&~\ref{tab:dataset_2}, researchers increasingly adopted deeper and more powerful CNN architectures. Backbone networks such as ResNet, Xception and so on~\cite{he2016deep,chollet2017xception,chen2017deeplab,chen2018encoder} emerged as standard backbones in crack segmentation models~\cite{lau2020automated,hsieh2021dau}.

Attention mechanisms were later integrated into CNN-based architectures to further refine feature representations and enhance localization~\cite{hsieh2021dau,qu2021crack,kang2022efficient,tao2023convolutional,wang2024dual}. For instance, DAU-Net~\cite{hsieh2021dau} introduces a channel attention block that reduces noisy activations and emphasizes salient features from the encoder, leading to more accurate crack segmentation. Similar designs have been proposed~\cite{qu2021crack,kang2022efficient,tao2023convolutional,wang2024dual}, confirming the effectiveness of attention in guiding the model’s focus toward crack regions.

Meanwhile, transformer-based backbones have been introduced into the domain~\cite{liu2021crackformer, guo2023pavement,tao2023convolutional,ding2023crack,xiang2023crack,jaziri2024hybrid,wang2024dual}. Unlike CNNs, which are prone to local features due to their kernel-based operations, transformers offer the advantage of capturing long-range dependencies and global context through attention mechanisms ~\cite{vaswani2017attention,dosovitskiy2020image}. This capability is particularly valuable for crack segmentation, where contextual information across distant parts of an image can help disambiguate challenging cases by differentiating between crack and non-crack features or other noise patterns~\cite{liu2021crackformer, guo2023pavement, ding2023crack}.

\textbf{Intricate Architectures and Loss Design.} 
Alongside backbone evolution, architectural design has also progressed. Early CNN-based methods often used simple, sequential stacks of convolutional layers~\cite{zhang2017automated,zhang2018deep} (Fig.~\ref{fig:sl} (a)).

Over time, encoder-decoder structures has become dominant~\cite{choi2019sddnet,kang2022efficient,hsieh2021dau,wang2024dual}, as shown in Fig.~\ref{fig:sl} (b) \& (c). These structures, inspired by U-Net~\cite{ronneberger2015u} have been found to be effective for crack segmentation tasks. The encoder extracts high-level feature representations through successive downsampling, while the decoder progressively upsamples and reconstructs the spatial resolution to produce detailed segmentation masks. This architecture, usually further adapted~\cite{zou2018deepcrack, zhang2018deep, hsieh2021dau,liu2021crackformer,tao2023convolutional} or paired with specific loss designs~\cite{kang2022efficient, li2021fast,pantoja2022topo,hsieh2023pavement} for crack segmentation, allows the model to preserve both low-level spatial details and high-level semantic context, which is crucial for accurately detecting narrow and disconnected cracks that are otherwise prone to being overlooked.

\textbf{Design for Implementation.} There is increasing emphasis on models suitable for deployment in real-world scenarios. This includes designing lightweight networks that focus on both efficiency and accuracy~\cite{choi2019sddnet,kang2022efficient,liu2024crackscf,liu2025scsegamba}. For example, SDDNet~\cite{choi2019sddnet} employs densely connected separable convolutions, achieving competitive performance with significantly fewer parameters, up to 88× fewer than comparable models, making it attractive for real-time field applications. A recent work, SCSegamba~\cite{liu2025scsegamba}, introduces a structure-aware vision network built on the state space model Mamba~\cite{gu2023mamba,zhang2024survey}, achieving state-of-the-art results with a lightweight design.

Another practical consideration is adaptation to varied data acquisition platforms, such as unmanned aerial vehicles (UAVs)~\cite{liu2022industrial,yao2024cracknex}. In~\cite{liu2022industrial}, for instance, an improved camera calibration method is proposed to compute the full-field scale of a UAV-mounted gimbal camera. The approach allows for convenient indexing of scale factors across different flight attitudes, eliminating the need for repeated recalibration and facilitating consistent measurements across scenes.

As the foundational paradigm in DL-based crack detection, supervised learning has driven the initial progress in this field. Its development—including advancements in areas such as backbone networks and architectural design, has laid a strong foundation and provides valuable reference for alternative learning paradigms. However, one of its key limitations is the heavy reliance on large volumes of labeled data, which can be costly and time-consuming to obtain~\cite{mo2022review, zhu2023survey}. As a result, there is a growing trend in the community to explore more data-efficient approaches, such as semi-supervised learning, weakly supervised learning, domain adaptation, and few-shot learning. These methods aim to leverage the strengths of supervised learning while reducing its dependency on extensive manual annotation.

\subsection{Semi-Supervised Learning}

\begin{figure*}
    \centering
    \includegraphics[width=0.9\linewidth]{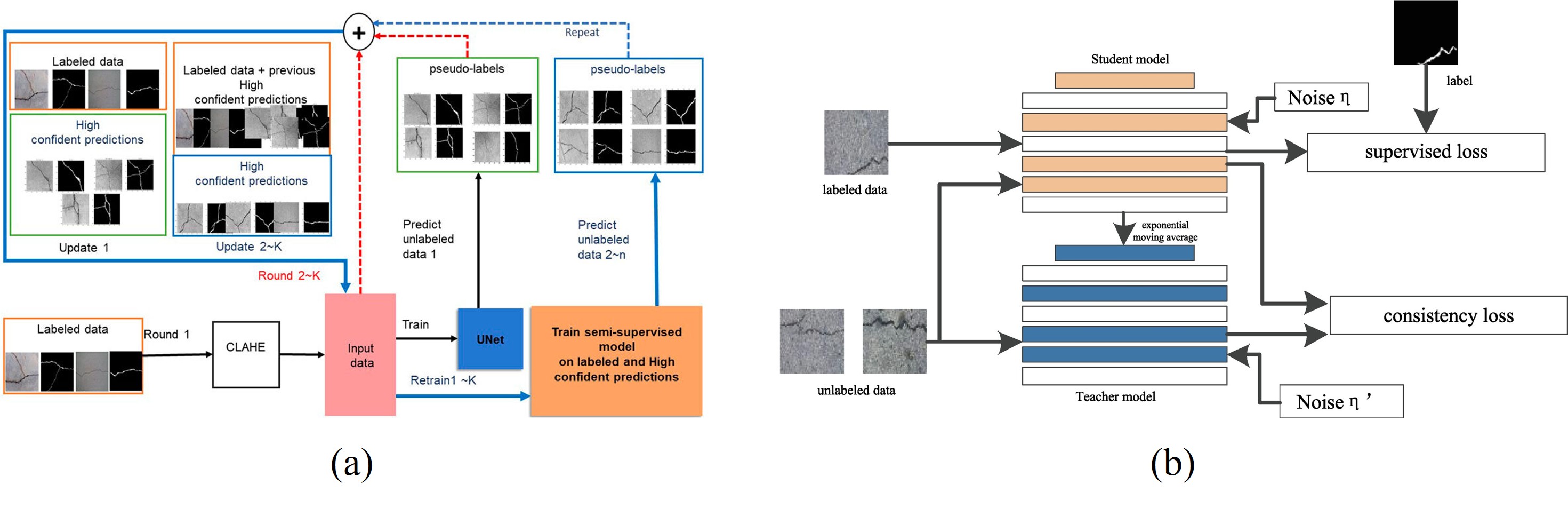}
    \caption{Representative works in semi-supervised crack segmentation, (a) for pseudo-labeling~\cite{mohammed2022end}, (b) for consistency regularization~\cite{wang2021semi}. }
    \label{fig:ssl}
\end{figure*}

Semi-supervised learning (SSL) in crack segmentation has begun to gain attention as researchers aim to reduce the dependence on the quantity of expensive pixel-level annotations while maintaining high segmentation performance.

In SSL, there is a subset of fully labeled samples $\mathcal{D}_L = \{(x_1, y_1), \ldots, (x_l, y_l)\}$, paired with another subset of unlabeled samples $\mathcal{D}_U = \{x_{l+1}, \ldots, x_n\}$. Usually $l \ll n$, meaning that the unannotated data makes up the majority of the whole dataset. The goal is to exploit both sets of data in order to train a model with better performance than one trained only with labeled data. This is especially important in real-world applications, as manual labeling of cracks is time-consuming and labor-intensive. As shown in Fig.~\ref{fig:ssl}, commonly used approaches in this domain can be categorized into two main trends: 1) pseudo-labeling, and 2) consistency regularization. Notably, another prominent trend on top of these individual methods is their combination to achieve improved performance.

\textbf{Pseudo-Labeling.} Pseudo-labeling, also known as self-training, is a widely adopted technique in semi-supervised learning~\cite{pelaez2023survey,yuan2021simple,he2021re,yang2022st++} and has shown noticeable promise in crack detection tasks~\cite{mohammed2022end}. The core idea is to leverage a small set of labeled crack images to train an initial supervised model, which is then used to generate predictions on unlabeled data. Based on carefully chosen criteria such as confidence level~\cite{mohammed2022end}, decision by a discriminative network~\cite{hung2018adversarial}, etc., predictions that meet the requirements will be selected as pseudo labels and then be incorporated into the training set as additional labeled data for further fine-tuning~\cite{zhu2021improving,feng2022dmt}.

For example, the authors of~\cite{mohammed2022end} adopt this pseudo-labeling strategy for crack segmentation by leveraging a modified U-Net~\cite{ronneberger2015u} architecture as the core segmentation network. The training process begins in a supervised manner, where the model learns from a small set of annotated crack images. Once the initial model is trained, it is used to generate predictions on a larger set of unlabeled data. The predicted probabilities are taken as confidence scores, with only predictions with higher scores than a threshold added to the original training set. After the model is retrained on the augmented set iteratively, the semi-supervised model achieves competitive accuracies compared to supervised models with only 40\% of annotated data. 

In~\cite{li2020semi}, authors first train a segmentation network in a fully supervised way, and also train a discriminator to distinguish ground truth from prediction. Then they uses the trained segmentation network to predict on the unlabeled data. The trained discriminator can provide a confidence score, which can determine if the prediction is good. If so, it will be used in the cross-entropy loss. Similarly, in~\cite{shim2020multiscale}, authors apply the idea of adversarial learning to train a discriminator to provide a confidence map for predictions on unlabeled data. They further use multi-scale features to enhance the performance.

Another group of works explore hybrid approaches combining multiple approaches, including pseudo-labeling, which will be discussed in the combined methods below.

\textbf{Consistency Regularization.} Consistency regularization is a fundamental principle in semi-supervised learning that has gained traction in crack detection tasks due to its simplicity and effectiveness~\cite{pelaez2023survey,tarvainen2017mean}. The central idea is that a model’s predictions should remain stable or consistent when the input image is subject to small perturbations or separate models are utilized for robust performance to pursue the same prediction goal.

In~\cite{wang2021semi}, authors use EfficentNet~\cite{tan2019efficientnet} to extract multi-scale crack feature information for crack segmentation. Two models are initialized as the student and teacher network with the same structure. The weights in the student network are updated directly using gradient descent, while the teacher network is updated by exponential moving average weights of the student model. Additionally, noise is added to the input data, and the student and teacher network are regularized by the consistency loss on the outputs.

In~\cite{guo2024pavement}, it integrates fractal dimension analysis~\cite{wu2020effective}, which is a method used to characterize how a structure's complexity changes across different scales, and semi-supervised learning. They first use fractal dimension analysis to preliminarily determine the candidate crack regions, then use the Crack Similarity Learning Network (CrackSL-Net) to learn the semantic similarity of crack image regions as in SimCLR~\cite{chen2020simple}, enhancing robustness and accuracy in semi-supervised settings.

\textbf{Combined Methods.} Another approach is to combine pseudo-labeling and consistency regularization for a further enhanced performance~\cite{jian2024cross,shamsabadi2024efficient}.

In~\cite{jian2024cross}, authors propose a cross-teacher-pseudo-supervision framework and cross-augmentation strategy. The proposed method employs two pairs of teacher–student models to mutually supervise each other using pseudo-labels generated from their respective teacher models. To boost the performance of the proposed algorithm, perturbations are applied to input, feature, and network during training. In~\cite{shamsabadi2024efficient}, the training process is divided into three sequential stages: 1) supervised training on labeled data, 2) consistency regularization to align predictions under input perturbations, and 3) pseudo-label generation based on prediction certainty.

Semi-supervised learning provides an efficient solution to reduce the dependence on large-scale fully annotated datasets. By leveraging both labeled and unlabeled data during training, this approach enhances model performance while minimizing the annotation burden. In the context of crack segmentation, semi-supervised learning has demonstrated competitive results compared to supervised learning~\cite{li2020semi,shim2020multiscale,mohammed2022end}, effectively bridging the gap between limited labeled data and the need for high-performing models.

\subsection{Weakly-Supervised Learning}

\begin{figure}
    \centering
    \includegraphics[width=1.0\linewidth]{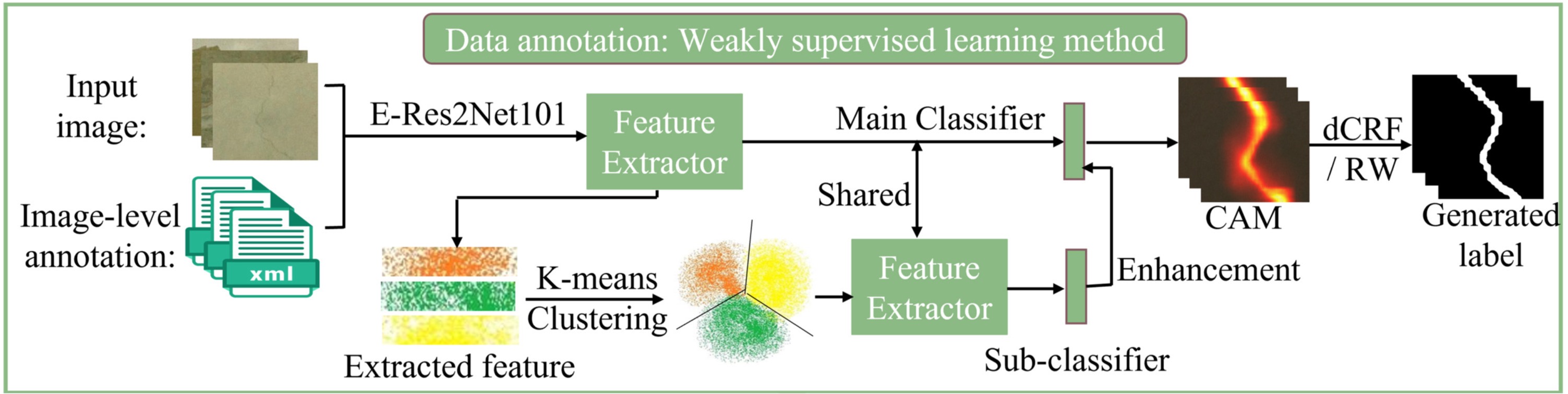}
    \caption{A representative work with image-level weak supervision~\cite{wang2021pixel}. Authors utilize binary image-level labels indicating the presence or absence of cracks to guide the weak supervision process.}
    \label{fig:wsl}
\end{figure}

Weakly-supervised learning has gained increasing attention in recent years as another data-efficient alternative to fully-supervised approaches. Weakly-supervised crack segmentation aims to alleviate this hardship by training models using weaker forms of annotation, including image-level labels, bounding boxes, scribble (rough line) annotations, etc. A representative work with image-level weak supervision is shown in Fig.~\ref{fig:wsl}. Different from semi-supervised learning, weakly-supervised learning usually rely on rough annotation to alleviate annotation burden instead of using precise annotation of less quantity.

\textbf{Image-Level Annotations.} 
Image-level annotations are among the most commonly used weak supervision signals in crack detection and other domains~\cite{ahn2019weakly,yao2020saliency}, primarily due to their ease of acquisition compared to dense pixel-level labels. These annotations are typically employed in conjunction with Class Activation Map (CAM)-based techniques~\cite{zhou2016learning}, which aim to localize discriminative regions associated with the class of interest by a trained model using gradient statistics. In other words, CAM helps visualize which regions of the image the model focuses on when making its decision. Foundational techniques such as GradCAM~\cite{selvaraju2020grad} and so on~\cite{chattopadhay2018grad,wang2020score,fu2020axiom} form the backbone of many weakly supervised approaches in this domain.

In~\cite{wang2021pixel}, the authors utilize binary image-level labels indicating the presence or absence of cracks to guide the weak supervision process. They introduce a method named Crack-CAM, which first generates coarse pixel-level maps using CAM, and then refines these maps with the help of Dense Conditional Random Fields (DenseCRF)~\cite{krahenbuhl2011efficient} and the Random Walk (RW) algorithm~\cite{xia2019random}. The refined labels are subsequently used to train a segmentation model in a fully supervised manner, effectively bridging the gap between weak supervision and full supervision.

Building upon a similar idea, Huangfu et al.~\cite{huangfu2025unified} propose a comprehensive refinement strategy known as AT-CAM (Affine Transformation and Pseudo Label Refinement). Their approach integrates XGradCAM~\cite{fu2020axiom} with geometric data augmentation through affine transformations. It further incorporates dual-hook denoising and dynamic range compression mechanisms to suppress noise in the activation maps and emphasize the structural characteristics of cracks. This framework achieves a 7.2\% relative improvement in segmentation accuracy compared to the baseline.

An alternative strategy is introduced in~\cite{dong2020patch}, where the authors reduce annotation costs by annotating image patches rather than entire images or pixels. A classification model is trained on these patch-level labels, and the resulting CAMs are used to infer pixel-level supervision. To improve the spatial coherence of the generated labels, DenseCRF~\cite{krahenbuhl2011efficient} is applied. This patch-based labeling strategy significantly reduces the annotation burden, by approximately 80\%, while still enabling the training of an effective segmentation model.

Works like~\cite{konig2022weakly, liu2023weakly, wang2024weakly, mishra2024weakly, jiang2024weakly, liang2025crackclip} also apply this idea of image-level or patch-level annotation in their pixel-level segmentation tasks. These methods collectively demonstrate the potential of combining CAM-based localization, refinement algorithms, and minimal supervision to achieve high-quality crack segmentation, even in the absence of detailed pixel-level annotations.

\begin{figure*}
    \centering
    \includegraphics[width=0.8\linewidth]{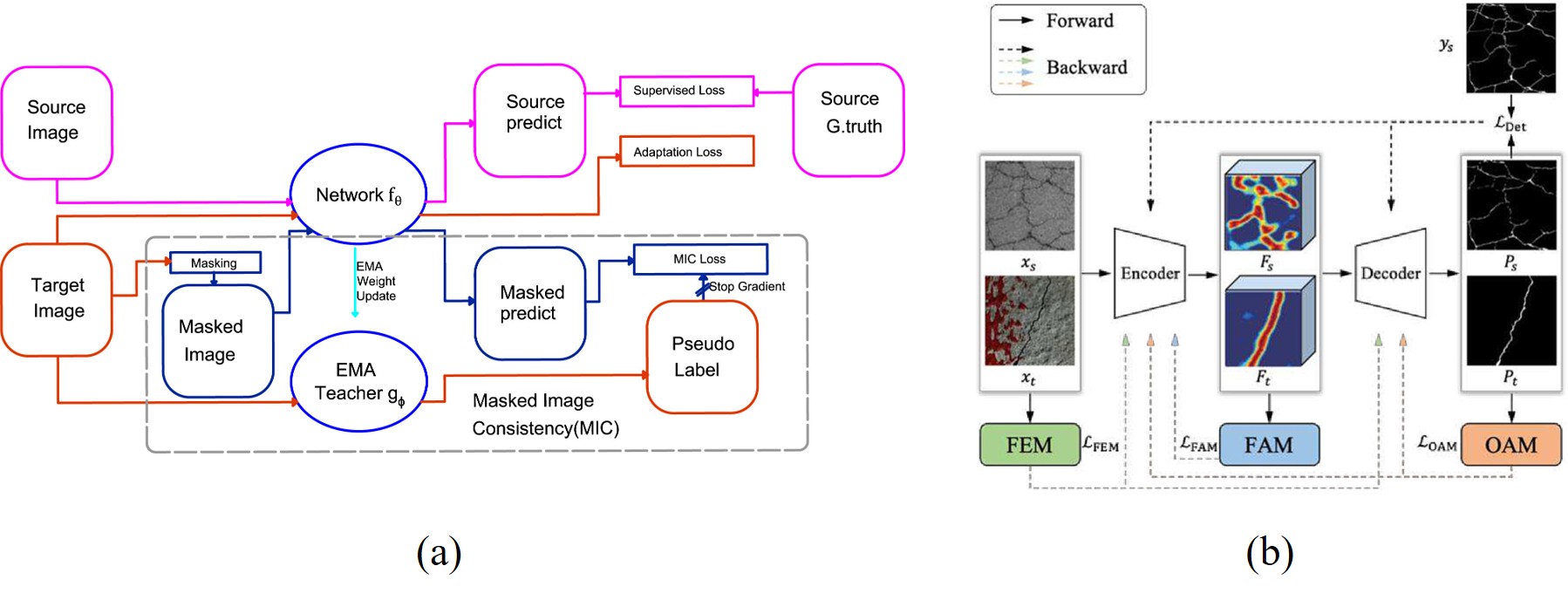}
    \caption{Representative works in domain adaptation-based crack segmentation, (a) for pseudo-labeling~\cite{beyene2023unsupervised}, (b) for domain alignment~\cite{weng2023unsupervised}.}
    \label{fig:da}
\end{figure*}

\textbf{Other Rough Annotations.} 
In addition to image- and patch-level annotations, bounding-box supervision and coarse pixel-level labeling such as scribble and keypoint also offer an intermediate form of supervision that balances annotation effort and ground-truth guidance, because of which, crack detection and other tasks~\cite{song2019box,oh2021background,lin2016scribblesup,liang2022tree} have begun to adopt such weak annotations. These forms of weak supervision enable the derivation of pixel-level labels using additional priors or image processing strategies.

In~\cite{zhang2022investigation, li2024implementation}, the authors explore bounding box-based supervision for crack segmentation. In~\cite{li2024implementation}, to convert coarse bounding boxes into pixel-level pseudo labels, they employ a multi-step refinement pipeline. Initially, background noise is suppressed, and high-confidence seed regions within the boxes are identified. Next, a region-growing algorithm is applied to expand the seed areas and generate rough segmentation maps. Finally, the GrabCut algorithm~\cite{rother2004grabcut} is employed to iteratively refine the segmentation, resulting in progressively improved pseudo labels. This method effectively bridges the gap between bounding-box annotations and full supervision without requiring costly per-pixel annotations.

Works like~\cite{inoue2021crack, tang2022weakly} use rough annotation instead. Tang et al.~\cite{tang2022weakly} proposes a method based on roughly annotated crack and non-crack regions. Instead of relying on precise contours, they utilize approximate labeling to extract three distinct types of supervision signals: 1) crack pixel labels, representing pixels that must be classified as crack; 2) non-crack pixel labels, indicating pixels that must be labeled as background; and 3) isolated pixel masks, which encode uncertain but informative regions that may contribute additional contextual cues. These labels are derived using classical image processing techniques, allowing the network to be trained with both strong and weak supervisory signals. This hybrid approach effectively balances annotation effort with segmentation performance.

By reducing the dependency on exhaustive manual labeling, weakly-supervised crack segmentation facilitates the deployment of segmentation models in real-world applications where annotation resources are limited, and it opens up new opportunities for leveraging originally unlabeled image collections. Nonetheless, achieving performance comparable to fully-supervised methods remains a significant research challenge, driving ongoing efforts to design more effective learning frameworks, loss functions, and regularization strategies tailored to weak supervision~\cite{dong2020patch,zhang2022investigation,jiang2024weakly}.

\subsection{Domain Adaptation}

The trend of domain adaptation (DA) in crack segmentation has become increasingly prominent, driven by the need for models that can generalize across diverse imaging conditions, pavement types, and sensor setups without requiring extensive retraining or annotations for each new domain.
In this setting, the source domain dataset is fully annotated, while the target domain typically has few or no annotations. Crack images collected from different regions, under varying lighting conditions, materials, and camera setups, often exhibit significant appearance differences, making domain shift a critical challenge. However, this scenario is highly practical where real-world data differs from annotated data because of diverse conditions, making it a meaningful direction for exploration.
Similar to SSL, there have emerged two main groups of methods: 1) pseudo-labeling and 2) domain alignment, as shown in Fig.~\ref{fig:da}.

\textbf{Pseudo-Labeling.} 
It has been a widely adopted strategy for domain adaptation in crack detection and other semantic segmentation tasks~\cite{tranheden2021dacs,hoyer2022daformer,hoyer2023mic}. It is also referred to as self-training in the literature~\cite{liu2022deep}. The core idea is to generate plausible pseudo labels for the target domain by leveraging the information learned from the source domain, enabling the segmentation network to adapt and perform better on unseen target domain data without requiring ground-truth annotations. This mechanism facilitates the transfer of knowledge and alleviates the domain shift problem that often hampers generalization across different datasets.

In~\cite{beyene2023unsupervised}, the authors demonstrate that pseudo-labeling-based approaches are effective for unsupervised domain adaptation in crack detection. Moreover, by incorporating masked image consistency into the DAFormer framework~\cite{hoyer2022daformer}, additional performance gains are achieved, emphasizing the importance of consistency regularization when adapting to new domains. Similarly, STDASeg~\cite{du2025self} utilizes an image blending-based domain mixing module to realize pseudo-labeling, as a self-training pipeline by the mutual learning scheme between CNNs and Transformers. A related concept is explored in~\cite{yu2023multi}, where the authors propose multi-source ensembled labels to assist in domain adaptation. Other studies enhance this mechanism using domain alignment techniques~\cite{chen2023deep,chun2024self}, as discussed below.

\begin{figure*}
    \centering
    \includegraphics[width=1\linewidth]{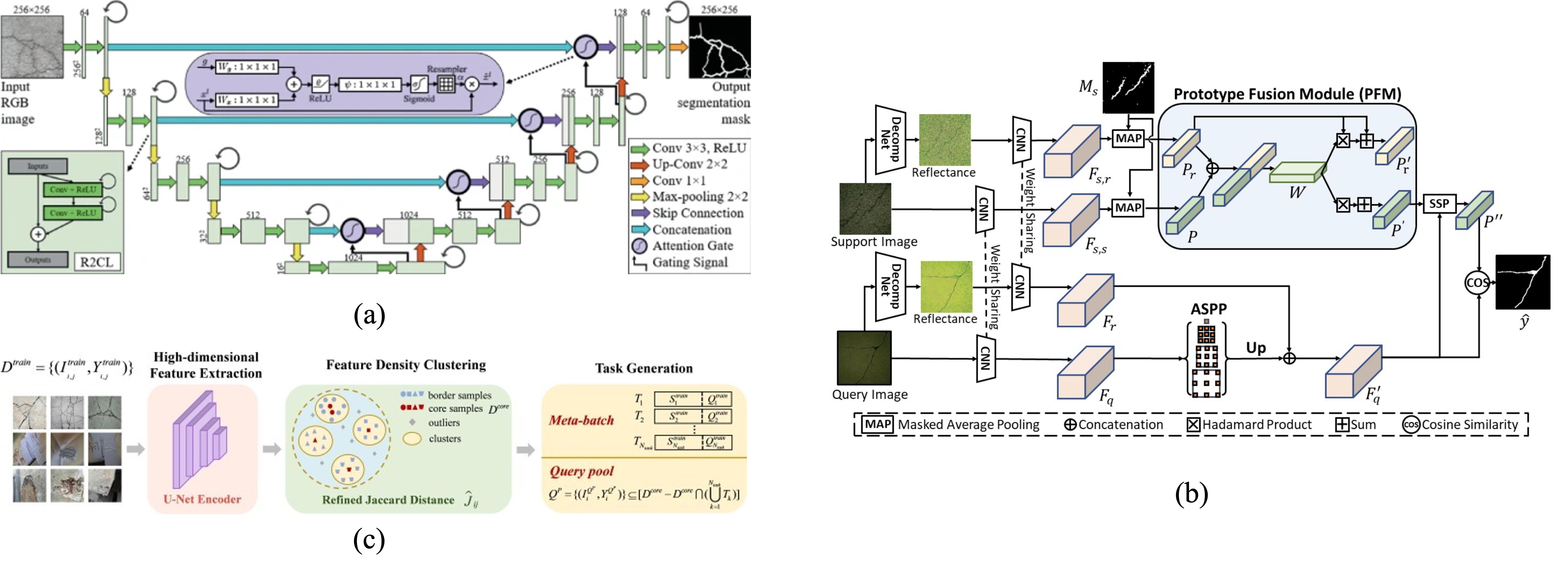}
    \caption{Representative works in crack segmentation based on few-shot learning, (a) for transfer learning~\cite{katsamenis2023few}, (b) for metric learning~\cite{yao2024cracknex}, (c) for meta learning~\cite{xu2023task}.}
    \label{fig:fsl}
\end{figure*}

\textbf{Domain Alignment.} 
Domain alignment is another widely adopted technique in domain adaptation for crack detection and segmentation. The core idea is to reduce the domain gap by aligning the source and target domains at different levels, typically the input space, feature space, or output space , so that a model trained on the source domain can generalize better to the target domain without being heavily affected by distribution shifts between domains.

In~\cite{weng2023unsupervised}, the authors propose DACrack, a comprehensive framework that performs domain alignment at three levels: input, feature, and output. To address input-level discrepancies, they design the Foreground Enhancement Module (FEM), which swaps low-level visual features between the source and target domains. This swapping process encourages the model to focus on structural similarities related to cracks while minimizing distractions from irrelevant textures, effectively serving as a form of contrastive attention. At the feature level, the Feature Adaptation Module (FAM) employs adversarial learning to align intermediate feature distributions across domains, helping the model learn domain-invariant representations. Finally, in the output space, they introduce the Output Adaptation Module (OAM), which utilizes a Variational Autoencoder (VAE)~\cite{kingma2013auto} to model and align the clustered manifold structure of cracks, ensuring the output distributions are consistent across domains.

Beyond pixel appearance, auxiliary signals such as depth have also been explored to bridge domain gaps. In~\cite{liu2022industrial}, the authors propose using disparity in depth prediction as a novel indicator of domain discrepancy. By estimating the inconsistency of predicted depth maps between the source and target domains, they introduce a depth-guided regularization term into the optimization process. This approach highlights the potential of leveraging additional modalities beyond RGB images, suggesting that domain adaptation can be strengthened by incorporating richer, complementary information.

Generative Adversarial Networks (GANs)\cite{goodfellow2014generative} have also found broad application in domain adaptation tasks. GANs are designed to learn a generative model that can map between different data distributions through adversarial training. A notable subclass, CycleGAN\cite{zhu2017unpaired}, enables image-to-image translation without requiring paired samples, making it especially valuable for unpaired domain adaptation. For example, in~\cite{zhang2025generative}, CycleGAN is utilized to merge features from shadow-free and shadowed datasets, effectively creating an adapted dataset with domain-invariant characteristics that facilitate crack detection under varying illumination conditions. Similarly,~\cite{zhao2025cross} employs CycleGAN for feature-level alignment without the need for annotations in the target domain. Their study investigates transfer directions and finds that bidirectional transfer, where both source and target features are mapped to a shared auxiliary space, better enforces an independent and identically distributed (i.i.d.) structure, mitigating risks of over-adaptation and preserving critical features throughout hierarchical network layers.

Together, these works highlight the growing sophistication of domain alignment techniques, evolving from simple adversarial feature matching to multi-level, multi-modality, and structure-aware adaptation strategies.

\subsection{Few Shot Learning}

Few-shot learning is another learning paradigm where a model is trained to generalize to new tasks using only a small number of labeled examples. This is particularly useful in crack detection where collecting large-scale annotated data is difficult or expensive~\cite{yang2024few,catalano2023few}. A typical problem formulation is N-way K-shot, where a model is tasked to perform N-class classification or segmentation given only K examples per class. In this domain, representative works can be categorized into three groups, 1) transfer learning, 2) metric learning and 3) meta learning, as shown in Fig.~\ref{fig:fsl}.

\begin{figure*}
    \centering
    \includegraphics[width=1\linewidth]{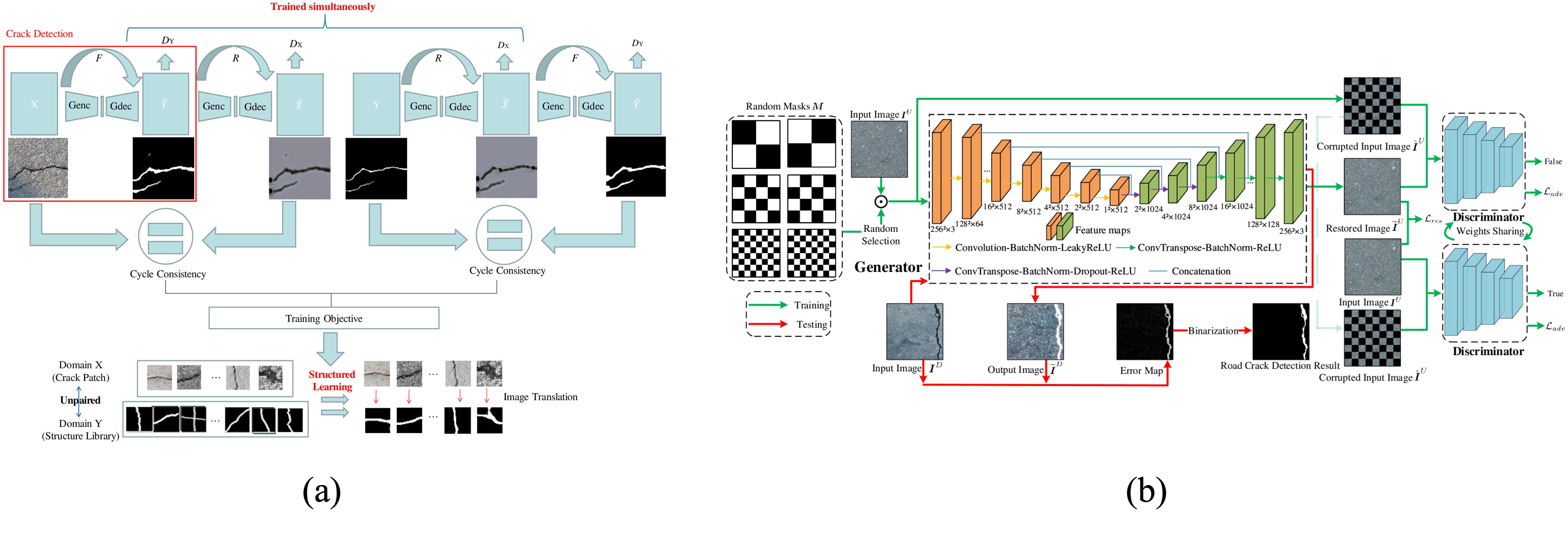}
    \caption{Representative works in crack segmentation based on unsupervised learning, (a) for detection by generation~\cite{zhang2020self}, (b) for anomaly detection~\cite{ma2024up}}
    \label{fig:usl}
\end{figure*}

\textbf{Transfer Learning.}
In the context of crack segmentation, few-shot learning is gaining attention due to the high cost and effort of pixel-level annotation. Transfer learning has been a widely used technique in deep learning-based crack detection~\cite{hsieh2021dau,8960998}. In the context of few shot crack detection, a model can be pre-trained on a large dataset from a different but related domain and fine-tuned on a small crack dataset~\cite{huang2021recovering,katsamenis2023few}. For instance, R2AU-Net~\cite{katsamenis2023few} leverages transfer learning and user feedback for few-shot adaptation. It dynamically refines model weights through incremental fine-tuning using manually rectified examples.

\textbf{Metric Learning.}
To better exploit the few-shot setup, metric-based approaches have gained attention. These methods compare feature representations between support and query sets using distance metrics~\cite{yang2024few,catalano2023few,snell2017prototypical}. In crack segmentation, this is achieved by extracting feature volumes from both sets and performing segmentation through similarity comparisons. CrackNex~\cite{yao2024cracknex}, for example, introduces a prototype-based framework where both support and reflectance prototypes are extracted and fused. The final segmentation is derived using cosine similarity between query features and these prototypes~\cite{fan2024cycle,feng2024suspected}.

\textbf{Meta Learning.}
On the other hand, meta-learning, often described as "learning to learn", aims to train models that can rapidly adapt to new tasks with minimal data. In crack segmentation, meta-learning frameworks typically involve episodic training on multiple segmentation tasks to mimic few-shot conditions. This way, the model learns a general strategy for quickly adapting to new crack segmentation problems, even when given very limited annotations~\cite{yang2024few,finn2017model}. This enables the model to acquire a generalizable adaptation strategy for crack segmentation~\cite{xue2023adaptive,xu2023task,zhong2024multi}. For example, in~\cite{xu2023task}, the authors propose a task generation method based on feature density clustering, selecting core samples to form the query pool before applying Model-Agnostic Meta-Learning (MAML)~\cite{finn2017model}, which improves interpretability and effectiveness of meta-training.

The application of few-shot learning in crack detection helps reduce the reliance on large training datasets while still maintaining reasonable accuracy and robustness, making it a practical and scalable solution for real-world scenarios.

\subsection{Unsupervised Learning}

Unsupervised learning aims to explore large collections of unlabeled data to uncover latent patterns, underlying structures, or meaningful representations without relying on explicit annotations or human-provided labels. In the context of crack segmentation, unsupervised approaches are particularly attractive because manual labeling of cracks is labor-intensive, subjective, and possibly prone to inconsistencies. There have emerged two main groups of methods, 1) detection by generation and 2) anomaly detection, as shown in Fig.~\ref{fig:usl}.

Earlier works have shown feasibility in unsupervised crack segmentation without deep learning. They usually rely on clustering methods~\cite{mubashshira2020unsupervised,oliveira2012automatic} or traditional image processing techniques~\cite{xu2013pavement} to distinguish crack from non-crack areas.

\textbf{Detection by Generation.} In deep learning-based unsupervised crack segmentation, generative models have been explored to learn the pixel-to-pixel translation between input crack images and segmentation maps, trained only with unpaired datasets, where crack images are not paired with dedicated crack maps~\cite{zhang2020self,duan2020unsupervised}. In ~\cite{zhang2020self}, a network is built with two GANs: one learns to translate a crack image patch to structured curves to represent cracks, and the other learns to conduct the reverse translation. Both GANs are trained simultaneously and regularized by cycle consistency, similar to CycleGAN~\cite{zhu2017unpaired}. It has been shown that the dual network can be trained to translate a cracked image to a ground truth (GT)-like image with a similar structural pattern, and it can be used for crack detection directly.

\textbf{Anomaly Detection.} On the other hand, methods such as~\cite{ma2024up,chow2020anomaly} adopt an anomaly detection approach, in which crack areas are the anomaly to be detected. For example, UP-CrackNet~\cite{ma2024up} first prepares randomly masked training images based on undamaged pavement images, then trains a generative network to restore missing regions with intact pavement patterns conditioned on masked images. During inference, the input image is masked similarly with a set of multi-scale square masks, and an error map is created by computing the pixel-wise difference between the input image and its reconstructed counterparts. This error map highlights cracks, as the generative network has learned to restore only undamaged areas.

Unsupervised learning is another practical approach in crack segmentation, particularly because it completely eliminates the need for manual annotations. However, note that despite removing the annotation burden, unsupervised learning still requires thoughtful data preparation. As research progresses, unsupervised learning continues to evolve as a promising path toward scalable and annotation-free crack detection.

\subsection{Foundation Models \& PEFT}

Generalization is becoming an increasingly important topic in the broader field of computer vision and has recently gained notable attention in the crack segmentation domain. The goal of generalization is to develop models that are not only highly accurate on data similar to training set but also robust when exposed to diverse real-world conditions, such as different pavement textures, lighting variations, imaging devices, and geographical locations. This growing focus is driven by the recognition that deep learning models trained on narrow or homogeneous datasets often suffer significant performance degradation when applied to unseen environments. As a result, researchers have begun to explicitly design crack segmentation methods with generalizability as a key objective, signaling a paradigm shift from purely performance-driven models to models that prioritize robustness and adaptability.

\begin{figure}
    \centering
    \includegraphics[width=1.0\linewidth]{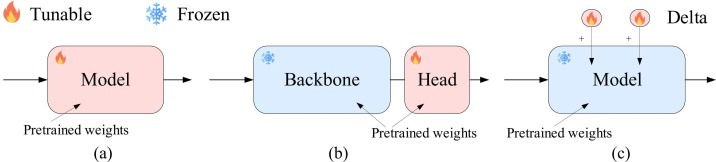}
    \caption{A recent representative work for generalization in crack segmentation~\cite{ge2024fine} based on parameter-efficient fine-tuing SAM~\cite{kirillov2023segment}.}
    \label{fig:generalization}
\end{figure}

A prominent influence in this shift is the emergence of vision foundation models represented by Segment Anything Model (SAM)~\cite{kirillov2023segment} which is trained on a large-scale dataset SA-1B (composed of of 11 million images and 1.1 billion masks) and capable of general-purpose image segmentation in a zero-shot manner. SAM has been adopted and further fine-tuned for some professional domains and showed success, such as medical image segmentation~\cite{wu2025medical,ma2024segment,zhu2024medical}, as well as crack detection~\cite{ge2024fine, guo2024segment, luo2025multiscenario}. In~\cite{ge2024fine}, the authors propose CrackSAM, where SAM is fine-tuned using two parameter-efficient fine-tuning methods: adapters and Low-Rank Adaptation (LoRA)~\cite{hu2022lora}, as shown in Fig.~\ref{fig:generalization}. Similarly,~\cite{guo2024segment} employs LoRA to fine-tune SAM, demonstrating that this strategy can lead to significant improvements in segmentation accuracy across eight diverse crack datasets when compared to conventional crack segmentation models.

Expanding on this trend,~\cite{ye2024sam} introduces the MCrack1300 dataset, targeting masonry defect detection. The authors propose two novel, fully automated methods based on SAM. Both methods involve fine-tuning SAM’s encoder using LoRA: one connects the encoder to alternative decoders, while the other utilizes a learnable self-generating prompter to guide segmentation. Both strategies achieve notable improvements over baselines, showcasing the versatility of prompt-based fine-tuning.

In~\cite{rostami2025segment}, a selective fine-tuning strategy is introduced, where only normalization parameters within the SAM architecture are fine-tuned. This selective tuning not only outperforms full fine-tuning and other PEFT techniques in terms of generalization performance but also achieves higher computational efficiency, making it an appealing option for real-world deployment.

Further extending the application of SAM to crack detection on edge devices,~\cite{wang2024crack} proposes CrackESS, a comprehensive system for detecting and segmenting concrete cracks. CrackESS first uses a YOLOv8~\cite{yolov8} model to generate self-prompts to provide initial localization and then applies a LoRA-based fine-tuned SAM for segmentation. The resulting masks are further refined by a dedicated Crack Mask Refinement Module (CMRM), leading to improved segmentation precision.

Collectively, these pioneering works represent some of the earliest and most promising efforts to adapt and fine-tune SAM for crack segmentation. Their success highlights the potential of leveraging large-scale vision models and parameter-efficient tuning methods to tackle the challenges of generalization, setting a new research direction for crack detection under diverse and real-world conditions.

\subsection{Generalizability Analysis}

Supervised learning has long been the dominant paradigm in deep learning-based crack segmentation, serving as the foundational approach that enabled significant breakthroughs in this domain. While traditional supervised learning does not explicitly aim for generalizability across domains, it provides the critical training framework upon which most models are built. Moreover, the recent emergence of foundation models that are trained with supervised objectives on larger datasets, demonstrates that supervised learning remains a cornerstone, even in systems designed for broader generalization.

Building on this foundation, alternative learning paradigms have gained popularity, particularly due to their potential to enhance model generalizability and reduce the reliance on labor-intensive annotations to achieve data efficiency. Semi-supervised learning leverages a small set of labeled data alongside a larger pool of unlabeled data, enabling effective training even in scenarios with limited supervision. Weakly supervised learning further reduces annotation requirements by using coarse or indirect labels to guide the learning process. Few-shot learning aims to generalize from only a few labeled examples per class, which is particularly valuable in real-world scenarios where collecting a large annotated dataset is impractical.

Domain adaptation addresses the challenge of domain shift, where models trained on one dataset may perform poorly on another due to differing characteristics. It enables models to adapt from a labeled source domain to a different, often unlabeled, target domain by aligning distributions at the image, feature, or label level. Unsupervised learning represents an even more annotation-efficient strategy, where models learn underlying structures and patterns directly from unlabeled data, though often with limited task-specific accuracy unless combined with auxiliary supervision.

Finally, foundation models represent a synthesis of these trends. Pre-trained on massive and diverse datasets using supervised objectives, they possess strong generalization capabilities across domains. 

In crack segmentation, the adaptation of such models and learning paradigms offers a promising direction for data efficiency and generalizability in diverse and challenging environments.

\begin{table}
    \centering
    \caption{Summary of publicly available datasets for crack segmentation (Table 1/2).}
    \label{tab:dataset_1}
    \large
    \renewcommand\arraystretch{1.35}
    \resizebox{\columnwidth}{!}{%
    \begin{tabular}{ccp{8cm}p{3cm}}
    \toprule
         Dataset&  Year& Description &Image Characteristics\\
    \midrule
        CrackTree200~\cite{zou2012cracktree} & 2012 & 206 pavement images containing various types of cracks, including distractions such as shadows and occlusions. & 600$\times$800 \\

        CrackIT~\cite{oliveira2014crackit} & 2014 & 84 pavement surface images collected using an optical device during a traditional road survey. & 1536$\times$2048, 1 pixel $\approx$ 1 mm$^{2}$, grayscale \\
        
        CFD~\cite{shi2016automatic} & 2016 & 118 urban road crack images captured using a smartphone with fixed camera settings in Beijing, containing common noise such as shadows. & 320$\times$480 \\
        
        Crack500~\cite{zhang2016road} & 2016 & pavement images collected on the Temple University campus using a smartphone. & 2448$\times$3264, 3-channel (RGB) \\
        
        AEL~\cite{amhaz2016automatic} & 2016 & 269 pavement images (68 annotated) collected using five acquisition systems (AigleRN, ESAR, LCMS, LRIS, TEMPEST2). Also referred to as the Sylvie Chambon dataset. & No uniform size \\
        
        CSSC~\cite{yang2017deep} & 2017 & 954 concrete crack images collected via web search; crack shapes are stochastically distributed. & 100$\times$100 and 130$\times$130 \\
        
        FCN~\cite{yang2018automatic} & 2018 & 800+ images of pavement and concrete wall cracks with varying widths (1--100 pixels) and complex shapes. Includes internet and real-world images from Harbin, China. & 224$\times$224; resolution: 72–300 dpi \\
        
        CRKWH100~\cite{zou2018deepcrack} & 2018 & 100 pavement images captured using a line-array camera under visible-light illumination. & 512$\times$512, ground sampling distance: 1 mm \\
        
        CrackTree260~\cite{zou2018deepcrack} & 2018 & 260 pavement images (expanded from CrackTree200~\cite{zou2012cracktree}) captured using an area-array camera under visible-light illumination. & 600$\times$800 \\
        
        CrackLS315~\cite{zou2018deepcrack} & 2018 & 315 pavement images captured using a line-array camera under laser illumination. & 512$\times$512, ground sampling distance: 1 mm \\
        
        Stone331~\cite{zou2018deepcrack} & 2018 & 331 stone surface images captured using an area-array camera under visible-light illumination. & 512$\times$512 \\
        
        DeepCrack~\cite{liu2019deepcrack}& 2019& 537 crack images from internet and real-world sources, covering varied textures, materials (asphalt, concrete), and crack widths (1--180 pixels). & 544$\times$384, 3-channel (RGB) \\
        
        GAPs384~\cite{yang2019feature}& 2019& 384 asphalt pavement crack images selected from the German Asphalt Pavement Distress (GAPs) dataset~\cite{eisenbach2017get}. & 1920$\times$1080, 1.2 mm/pixel, 8-bit grayscale\\
        
        KolektorSDD~\cite{tabernik2020segmentation}& 2019& 52 crack images contained in 399 electrical commutator surface images. & 1408$\times$512\\
        
        Khanh11k~\cite{Khanh}& 2019& 112,000 images with varied crack conditions and environments, merged from multiple datasets including CrackTree200~\cite{zou2012cracktree}, Crack500~\cite{zhang2016road}, GAPs384~\cite{yang2019feature}, CFD~\cite{shi2016automatic}, and AEL~\cite{amhaz2016automatic}. & 448$\times$448\\          
        UAV75~\cite{benz2019crack}& 2019& 75 images captured by unmanned aircraft systems (UAS), containing various cracks and planking patterns (visually similar to cracks). & 512$\times$512\\

        DIC~\cite{rezaie2020comparison}& 2020 & 530 image patches from 8 laboratory stone masonry wall specimens, captured for digital image correlation (DIC) under various loading levels and support conditions. & 256$\times$256\\
    \bottomrule

    \end{tabular}
    }
    
\end{table}        

\begin{table}
    \centering
    \caption{Summary of publicly available datasets for crack segmentation (Table 2/2).}
    \label{tab:dataset_2}
    \large
    \renewcommand\arraystretch{1.35}
    \resizebox{\columnwidth}{!}{%
    \begin{tabular}{ccp{8cm}p{3cm}}
    \toprule
         Dataset&  Year& Description &Image Characteristics\\
    \midrule
        Masonry~\cite{dais2021automatic}& 2021 & 11,491 image patches (4,057 with cracks) from 469 masonry surface photos with varied resolutions, crack types, and background noise. & 224$\times$224\\
        
        Ceramic~\cite{junior2021ceramic}& 2021 & 167 ceramic crack images with varied crack shapes, ceramic types, and imaging conditions (e.g., scale, angle, illumination).  & 256$\times$256, 3-channel (RGB)\\
        
        CrSpEE~\cite{bai2021detecting}& 2021 & 2,229 images with cracks and spalling on various material types and at different scales. & 147$\times$288 to 4600$\times$3070 \\
        
        BCL~\cite{ye2021structural}& 2021 & 11,000 image patches from 50+ in-service bridges: 5,769 non-steel crack, 2,036 steel crack, 3,195 noise; under varied weather, lighting, and imaging conditions. & 256$\times$256\\
        
        Conglomerate~\cite{bianchi2021corrosion,bianchi2022development} & 2021 & 10,995 crack images synthesized from nine public crack datasets~\cite{zou2012cracktree,shi2016automatic,zhang2016road,eisenbach2017get,liu2019deepcrack,yang2019feature} via normalization, noise removal, and resizing. & 512$\times$512\\
        
        LCW~\cite{bianchi2021lcw,bianchi2022development} & 2021 & 3,817 bridge inspection images with varied lighting, background materials, crack widths, and scales. Named as Labeled Cracks in the Wild (LCW) dataset. & 512$\times$512 \\
        
        Syncrack~\cite{rill2022syncrack}& 2022 & 600 synthetic crack images (expandable) generated under varying noise levels using the proposed crack generator. & 480$\times$320, 3-channel (RGB)\\
        
        TopoDS~\cite{pantoja2022topo}& 2022 & 692 stone masonry images expanded from DIC~\cite{rezaie2020comparison}, including cracks on lab specimens and real-world damaged buildings. & 256$\times$256\\
        
        CrackSeg9k~\cite{kulkarni2022crackseg9k}& 2022& 9,255 crack images refined from ten public crack datasets~\cite{zou2012cracktree,shi2016automatic,zhang2016road,liu2019deepcrack,yang2019feature,dais2021automatic,junior2021ceramic} via resizing, noise removal, distortion, and refinement. & 400$\times$400 \\

        S2DS~\cite{benz2022image}& 2022& 232 crack images from the 743-image Structural Defect Dataset (S2DS), collected at real inspection sites using various camera platforms.& 1024$\times$1024\\

        FIND~\cite{zhou2023deep}& 2023& 2,500 crack image patches from bridge deck and roadways obtained by a laser scanning device, containing surface elevation information. & 256$\times$256, image types: raw intensity, raw range, filtered range, and fused\\

        CrackMap~\cite{katsamenis2023few}& 2023& 120 road crack images collected using a vehicle-mounted RGB camera, representing typical road scenes. & 256$\times$256, 3-channel (RGB)\\

        Crack900~\cite{huang2024crackfusion,zhang2023crack900} & 2023 & 914 sets of visible and infrared (IR) images captured from masonry walls using solar heating as the sole heat source. & 384$\times$288, image types: RGB, IR, RGB\_IR, RGB\_T, fused \\

        TUT~\cite{liu2024crackscf}& 2024& 1,408 crack images from mobile phones (1,270) and online sources (138), featuring diverse, real-world backgrounds and complex crack patterns. & 640$\times$640\\

        SegCODEBRIM~\cite{jaziri2024hybrid}& 2024 & 420 crack segmentation labels added to CODEBRIM, a concrete defect dataset for bridges. & 1500$\times$844\\

        MCrack1300~\cite{ye2024sam} & 2024 & 1,300 masonry crack images sourced from Crack900, online images, and mobile phone photos, covering diverse brick types and crack patterns. & 640$\times$640 \\
        
        OMNICRACK30K~\cite{benz2024omnicrack30k}& 2024& 30,017 crack images synthesized from 20 public datasets~\cite{zhang2016road,shi2016automatic,amhaz2016automatic,yang2017deep,zou2018deepcrack,benz2019crack,liu2019deepcrack,yang2019feature,Khanh,rezaie2020comparison,ye2021structural,bai2021detecting,dais2021automatic,junior2021ceramic,bianchi2021lcw,bianchi2022development,pantoja2022topo,benz2022image}, with duplicates and noise removed, and class imbalance addressed. & No uniform size \\
        
        \textbf{Ours}& 2025& 1,634 pavement 3D images from asphalt and concrete surfaces, collected using a mainstream survey system under varied environments and crack severities; each pixel encodes pavement elevation. & 512$\times$624, 4 mm/pixel, 8-bit grayscale \\
    \bottomrule

    \end{tabular}
    }
    
\end{table}

\section{Datasets}
\label{sec:datasets}

\subsection{Review}
With the rapid growth of the crack detection field, an increasing number of datasets have been introduced to support research and development. A timeline is summarized in Fig.~\ref{fig:combined}, showing releases of publicly available datasets based on different categories: 1) general crack segmentation dataset, 2) dataset with specialized sensors other than RGB cameras, 3) aggregated dataset combining existing datasets and 4) synthetic dataset. An exhaustive list of publicly available datasets is provided in Tab.~\ref{tab:dataset_1} \&~\ref{tab:dataset_2}.

Most publicly available datasets are collected using RGB cameras, as shown in the tables. These datasets typically include road or structure surfaces annotated with crack locations, and they have become foundational for benchmarking deep learning models.

Later, data collected using professional-grade sensors such as 3D laser scanners has gained popularity in real-world deployments due to their enhanced robustness under varying lighting and surface conditions. Early works~\cite{wang2011elements,wang2011automated,tsai2012critical,tsai2012pavement} began to utilize such data collected around 2012, and since then it has been widely adopted by researchers and practitioners~\cite{jiang2016enhanced,amhaz2016automatic,hsieh2021dau}. A survey reported
in 2017 shows that 18 US states use 3D automated data collection and 17 have a plan to use it in the plan~\cite{zimmerman2017pavement}. It has also been a well-recognized technology by the Federal Highway Administration (FHWA) and various state Departments of Transportation across the United States~\cite{FHWAInfoTechnology, pierce2019automated}. However, despite their value, datasets from such sensors are rarely shared publicly. Although datasets such as AEL~\cite{amhaz2016automatic} and FIND~\cite{zhou2023deep} provide annotated data collected with professional sensors, they fall short in quantity, resolution, or diversity, limiting their usefulness for DL models that need to be robust against distractions like pavement joints, markings etc. As a result, there remains a significant gap between academic research and practical implementation.

Moreover, datasets are not only increasing in number but also in scale. Newer datasets tend to contain thousands or even tens of thousands of annotated images by aggregating data from multiple sources~\cite{Khanh,kulkarni2022crackseg9k,benz2024omnicrack30k}, which enable the training of larger and more complex models to improve performance and generalize across multiple scenarios~\cite{ge2024fine}. 

At the same time, there has been a growing interest in synthetic datasets. These are generated using computer graphics, simulation, or generative models, and offer fine-grained control over crack characteristics, environmental conditions, and ground truth labels~\cite{zhai2022synthetic,rill2022syncrack}. Synthetic data can supplement real-world datasets by providing diverse, balanced, and fully labeled training samples, especially useful for tasks in which data is not easy to acquire or annotate.

\begin{figure}
    \centering
    \includegraphics[width=1\linewidth]{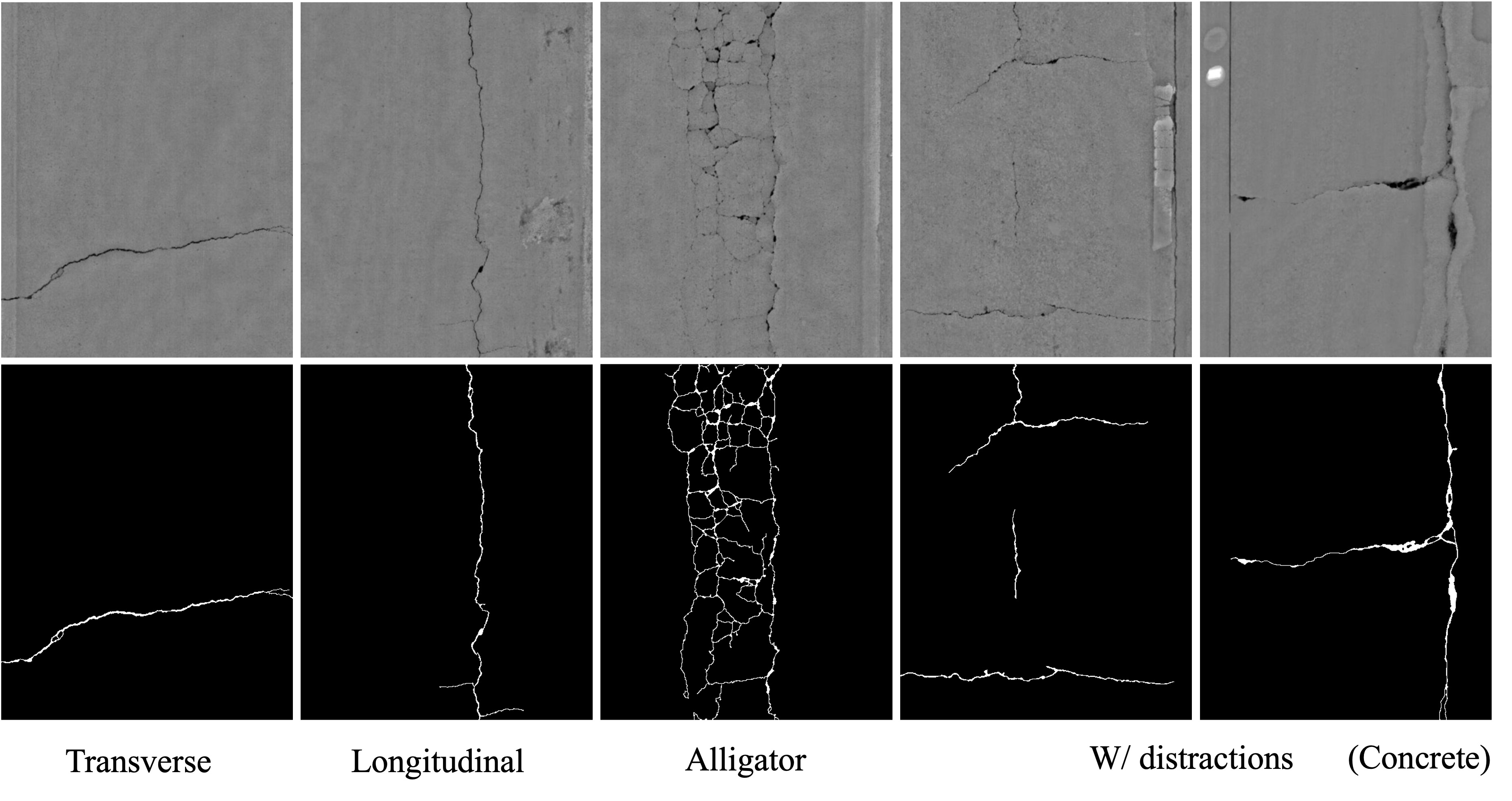}
    \caption{Examples from our \textbf{3DCrack} dataset. Each range image is paired with fine-grained pixel-level annotation. Cracks can be categorized into transverse, longitudinal, and compound (including alligator) cracks. Note that the last two examples show some common distractions, such as pavement marking and joints, which are not considered cracks. The last column includes an example of a concrete pavement surface, while others are flexible surfaces.}
    \label{fig:3dcrack}
\end{figure}

\subsection{A New Dataset: 3DCrack}

As introduced in Sec.~\ref{sec:datasets}, 3D laser scanning has become one of the mainstream technique for capturing high-resolution, full-lane pavement data, offering robust performance under varying lighting conditions. Unlike 2D intensity-based imaging, 3D laser systems are resilient to noise from shadows, stains, or low contrast lighting, making cracks and surface defects more distinguishable\cite{wang2011elements,wang2011automated,tsai2012critical,tsai2012pavement,jiang2016enhanced,amhaz2016automatic,hsieh2021dau}. Due to these advantages, 3D pavement data has been widely adopted by departments of transportation (DOTs) for real-world crack detection~\cite{zimmerman2017pavement,FHWAInfoTechnology}. However, the lack of large-scale, high-resolution datasets with common distractions remains a bottleneck for further research and implementation.

To bridge this gap, we release \textbf{3DCrack}, a new 3D pavement image dataset to support deep learning-based crack detection (Fig.~\ref{fig:3dcrack}). Data was collected using the Georgia Tech Sensing Vehicle (GTSV), equipped with dual 3D line laser sensors. These sensors capture a 4-meter-wide pavement profile at speeds up to 60 mph. Each frame produces a dense 1,000 × 2,080 point cloud, with 1 mm × 1 mm resolution after interpolation, and 0.5 mm height precision\cite{tsai2012critical,tsai2012pavement}. The point clouds are converted into 3D/range images via compression by rescaling height to 8-bit grayscale, with darker pixels representing depressions, and rectification by Gaussian high-pass filtering to suppress large-scale elevation trends~\cite{jiang2016enhanced,hsieh2020machine}.

The 3DCrack dataset includes 1,139 training, 245 validation, and 250 testing images collected from SR 275, US 80, and I-16 in Georgia, including both flexible and concrete pavement types. The dataset covers diverse cracking conditions, including varying geometries such as transverse, longitudinal, alligator, and compound cracks, as well as challenges like pavement joints and other visual distractions, ensuring robustness in trained models across real-world scenarios. Each image measures 512 × 624 pixels, representing 4-mm resolution per pixel. Ground-truth crack masks were manually annotated by trained annotators to ensure label quality as in~\cite{hsieh2020machine}.
\section{Experiments \& Findings}
\label{sec:experiments}

\begin{table}
\centering
\caption{Performance of Models on different datasets trained with different quantities. The \textbf{best performance} is highlighted in bold, and the \underline{second best} is underlined.}
\label{tab:exp1}

\resizebox{\columnwidth}{!}{%
\begin{tabular}{llcllcllccc}
\toprule
 \multirow{2}{*}{Method} & \multirow{2}{*}{Metric}  &\multicolumn{4}{c}{DeepCrack}&    &\multicolumn{4}{c}{3DCrack}\\ 
 \cmidrule(lr){3-6} \cmidrule(lr){8-11}
 &  & 25\%&  50\%& 75\%& 100\%&    &25\%&50\%& 75\%& 100\%\\ 
 
\midrule

 ResUNet&  \multirow{8}{*}{IoU}& {0.694}&  \underline{0.717}& \underline{0.725}&{0.725}&    &{0.456}&\textbf{0.648}& \textbf{0.694}& \textbf{0.703}\\
 DeepCrack&  & 0.667&  0.674& 0.701&0.706&    &0.435&0.504& \underline{0.639}& 0.662\\
 Deeplabv3+&  & 0.667&  {0.708}& {0.707}&0.718&    &0.385&{0.56}& 0.632& 0.635\\
 XcepUNet&  & 0.673&  0.684& 0.702&0.708&    &0.41&0.482& 0.622& 0.661\\
 SegFormer&  & 0.664&  0.701& 0.701&0.709&    &0.413&{0.56}& 0.632& 0.635\\
 UniMatch&  & 0.665&  0.672& 0.702&0.675&    &0.267&0.346& 0.551& 0.567\\
 UniMatchv2&  & \underline{0.696}&  0.700& \underline{0.707}&\underline{0.732}&    &\textbf{0.566}&0.47& 0.611& \underline{0.665}\\
 CrackSAM& & \textbf{0.723}& \textbf{0.748}& \textbf{0.748}& \textbf{0.751}& & \underline{0.564}& \underline{0.588}& 0.592&0.6\\

\midrule

 ResUNet&  \multirow{8}{*}{Precision}& 0.831&  \textbf{0.866}& 0.87&\underline{0.879}&    &{0.607}&\textbf{0.81}& \underline{0.82}& \textbf{0.827}\\
 DeepCrack&  & {0.838}&  0.834& \underline{0.873}&0.859&    &0.57&0.686& 0.762& \underline{0.805}\\
 Deeplabv3+&  & {0.838}&  \underline{0.839}& 0.855&0.834&    &0.52&\underline{0.705}& 0.736& 0.733\\
 XcepUNet&  & \underline{0.843}&  0.838& \textbf{0.874}&\textbf{0.895}&    &0.6&0.659& \textbf{0.827}& 0.803\\
 SegFormer&  & 0.795&  0.818& 0.844&0.829&    &0.53&\underline{0.705}& 0.736& 0.733\\
 UniMatch&  & 0.789&  0.819& 0.815&0.87&    &0.306&0.353& 0.608& 0.658\\
 UniMatchv2&  & 0.821&  0.784& 0.816&0.85&    &\textbf{0.736}&0.576& 0.694& 0.753\\
 CrackSAM& & \textbf{0.847}& 0.823& 0.849& 0.86& & \underline{0.661}& 0.685& 0.691&0.703\\

 \midrule

  ResUNet&  \multirow{8}{*}{Recall}& 0.825&  0.82& 0.827&0.817&    &0.552&\underline{0.727}& \textbf{0.778}& \underline{0.782}\\
 DeepCrack&  & 0.793&  0.799& 0.788&0.809&    &0.558&0.586& 0.738& 0.741\\
 Deeplabv3+&  & 0.779&  0.831& 0.817&\underline{0.853}&    &0.521&{0.683}& 0.763& 0.767\\
 XcepUNet&  & 0.788&  0.804& 0.787&0.783&    &0.486&0.566& 0.668& 0.739\\
 SegFormer&  & 0.822&  {0.839}& 0.82&0.844&    &0.563&{0.683}& 0.763& 0.767\\ 
 UniMatch&  & {0.837}&  0.811& {0.853}&0.767&    &{0.639}&0.654& 0.747& 0.725\\
 UniMatchv2&  & \underline{0.84}&  \underline{0.886}& \underline{0.863}&{0.852}&    &\underline{0.66}&0.626& \underline{0.765}& \textbf{0.811}\\
 CrackSAM& & \textbf{0.844}& \textbf{0.901}& \textbf{0.873}& \textbf{0.862}& & \textbf{0.716}& \textbf{0.736}& 0.745&0.748\\

 \midrule

   ResUNet&  \multirow{8}{*}{Dice}& {0.808}&  \underline{0.825}& \underline{0.833}&{0.833}&    &{0.557}&\textbf{0.747}& \textbf{0.788}& \textbf{0.795}\\
 DeepCrack&  & 0.784&  0.787& 0.81&0.812&    &0.538&0.602& 0.735& 0.758\\
 Deeplabv3+&  & 0.786&  {0.819}& 0.818&0.827&    &0.496&{0.673}& \underline{0.74}& 0.741\\
 XcepUNet&  & 0.79&  0.794& 0.811&0.817&    &0.503&0.582& 0.72& 0.754\\
 SegFormer&  & 0.785&  0.813& 0.816&0.821&    &0.521&{0.673}& \underline{0.74}& 0.741\\ 
 UniMatch&  & 0.783&  0.786& 0.815&0.793&    &0.361&0.434& 0.66& 0.675\\
 UniMatchv2&  & \underline{0.812}&  0.816& {0.822}&\underline{0.84}&    &\underline{0.673}&0.585& 0.72& \underline{0.771}\\
 CrackSAM& & \textbf{0.828}& \textbf{0.848}& \textbf{0.848}& \textbf{0.85}& & \textbf{0.676}& \underline{0.699}& 0.704&0.711\\

 \midrule

\end{tabular}
}

\end{table}

To facilitate future research, we benchmark several representative models on the 3DCrack and DeepCrack~\cite{liu2019deepcrack} datasets to establish baseline performance. Two experimental settings are considered. The first evaluates supervised learning methods under varying training data quantities, providing reference points not only for supervised learning but also for semi-supervised learning approaches. The second focuses on generalization performance, where models are trained on a large-scale set and tested on a distinct set with different surface conditions, offering baselines for generalizable crack detection methods.

To assess model performance, we adopt four commonly used metrics: Intersection over Union (IoU), Precision, Recall, and Dice coefficient. These metrics are defined as follows: TP is for True Positives, FP for False Positives, and FN for False Negatives.

\begin{equation} 
\text{IoU} = \frac{TP}{TP + FP + FN}
\end{equation}

\begin{equation} 
\text{Precision} = \frac{TP}{TP + FP}
\end{equation}

\begin{equation} 
\text{Recall} = \frac{TP}{TP + FN}
\end{equation}

\begin{equation} 
\text{Dice} = \frac{2 \cdot TP}{2 \cdot TP + FP + FN}
\end{equation}

In the following experiments, we conduct the same training and test setting on our 3DCrack and DeepCrack~\cite{liu2019deepcrack} datasets to quantify how well each model performs on different datasets and highlight the characteristics and challenges specific to our 3DCrack. The original DeepCrack does not have a validation set, therefore we split the original test set into validation and test set with a ratio of 1:1.

\subsection{Supervised \& Semi-Supervised Experiments}

\begin{figure*}
    \centering
    \includegraphics[width=1\linewidth]{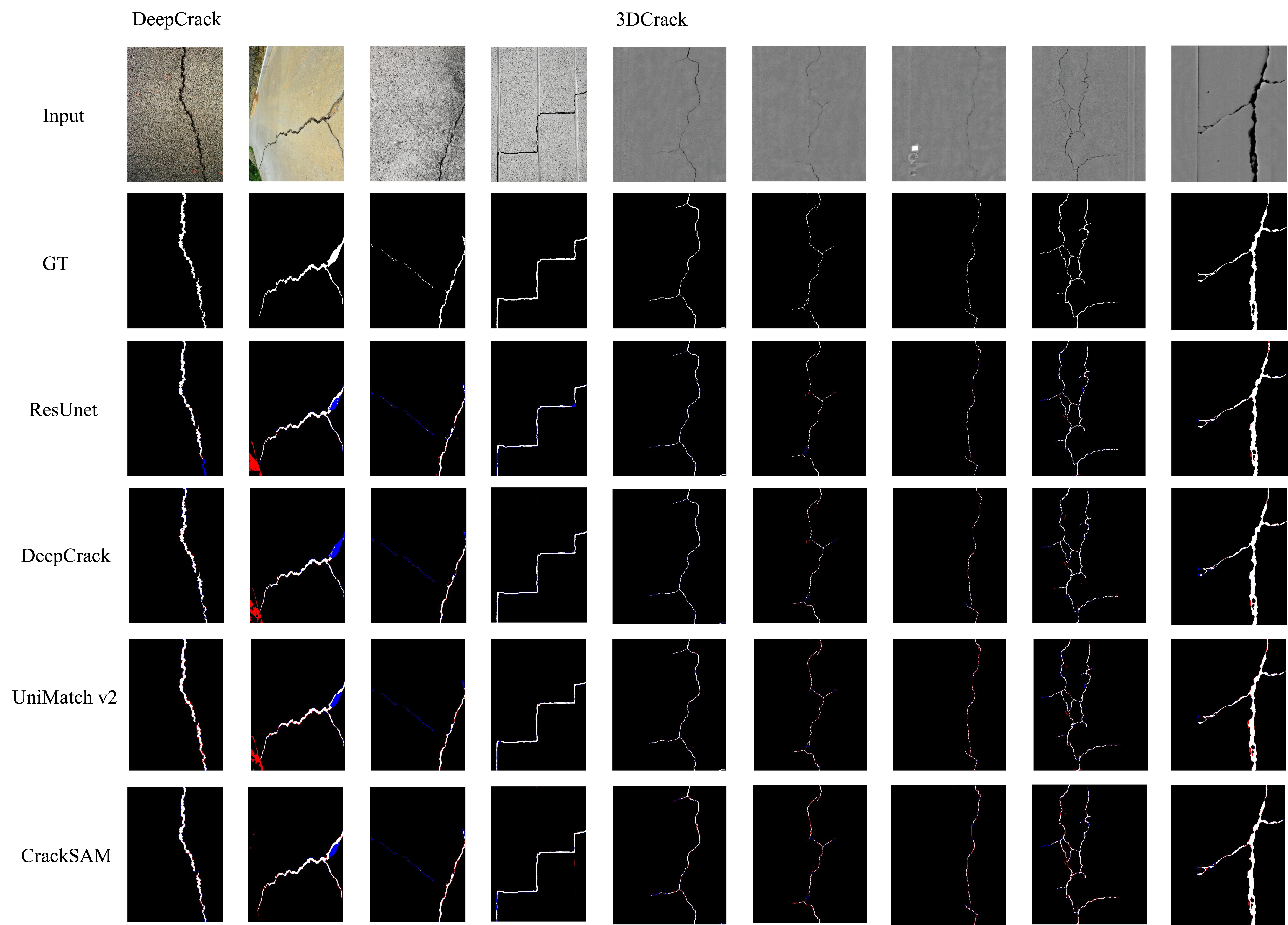}
    \caption{Crack segmentation results on DeepCrack and 3DCrack datasets from different models. False positives are highlighted in red, and false negatives are highlighted in blue.}
    \label{fig:exp1}
\end{figure*}

\subsubsection{Experimental Settings}

For the supervised experiments, we selected a set of representative models based on their proven effectiveness in image segmentation tasks and the relative ease with which they can be implemented and fine-tuned. Specifically, we include the U-Net ~\cite{ronneberger2015u} architecture with two distinct backbone encoders: ResNet ~\cite{he2016deep} and Xception~\cite{chollet2017xception}. The ResNet backbone is widely recognized for its residual learning capabilities, which help mitigate the vanishing gradient problem and improve feature propagation. The Xception backbone, on the other hand, leverages depthwise separable convolutions to enhance model efficiency and accuracy. Additionally, we incorporate DeepCrack~\cite{liu2019deepcrack}, a model explicitly tailored for crack detection, which has shown strong performance in identifying fine-grained linear structures in images. Deeplabv3+~\cite{chen2018encoder} is included for its robust semantic segmentation capabilities, particularly its use of atrous spatial pyramid pooling and encoder-decoder design that facilitates precise boundary localization. SegFormer~\cite{xie2021segformer} is also evaluated due to its strong performance in recent benchmarks, combining transformer-based encoding with lightweight decoders for efficient and accurate segmentation. A recent foundation model for crack segmentation, CrackSAM~\cite{ge2024fine} is also included here and will be detailed below.

For the semi-supervised experiments, the aforementioned models are adapted in a straightforward manner by limiting the number of labeled training samples, allowing us to evaluate their robustness under label-scarce conditions. To complement these baselines, we also include UniMatch v1\&v2 ~\cite{unimatch, unimatchv2}, a state-of-the-art model family designed specifically for semi-supervised semantic segmentation. UniMatch employs consistency regularization and pseudo-labeling strategies to leverage unlabeled data effectively and has demonstrated superior performance across several benchmarks under limited supervision. Moreover, UniMatch v2 employs DINOv2~\cite{oquab2023dinov2} as a more recent and powerful encoder.

Models are trained according to the original training strategies specified in their papers. Early stopping is applied during training, with a maximum of 100 epochs for each model.

\subsubsection{Findings}

\textbf{Overall Performance.} As presented in Tab.~\ref{tab:exp1} and Fig.~\ref{fig:exp1}, all selected models are capable of segmenting cracks effectively after training. Among them, ResUNet delivers a robust overall segmentation performance across both datasets when trained on the full training set, achieving a Dice score of 0.833 on DeepCrack and 0.795 on 3DCrack. The recent UniMatch v2 model also performs well, benefiting from its use of the strong DINOv2 vision encoder.

The two specialized crack detection models, DeepCrack and CrackSAM, exhibit notably different behaviors. Despite being specialized for this task, DeepCrack sightly lags behind models like ResUNet, likely due to its shallow and simple architecture. In contrast, CrackSAM achieves consistently high scores across metrics, training data quantities, and datasets. This highlights one fundamental development in DL-based crack detection: stronger backbones are beneficial to performance. CrackSAM inherits powerful semantic understanding from the Segment Anything Model (SAM)~\cite{kirillov2023segment}, enabling it to better distinguish cracks from common non-crack distractions. This advantage is illustrated by an example in the second column of Fig.~\ref{fig:exp1}, where CrackSAM more effectively suppresses false positives in non-crack regions compared with other models.

\textbf{Dataset Difference.} Across the board, models perform worse on the 3DCrack dataset than on DeepCrack, highlighting the domain differences between the two datasets. While DeepCrack contains common distractions such as shadows and non-crack areas, it generally provides clearer and more zoomed-in views of cracks. In contrast, 3DCrack contains finer, more subtle hairline cracks due to its downward-looking perspective and broader coverage of the pavement surface as shown by examples in Fig.~\ref{fig:3dcrack}. These cracks are often more difficult to detect, leading to a noticeable drop in performance. For instance, ResUNet's IoU score decreases from 0.725 on DeepCrack to 0.703 on 3DCrack using 100\% of the training data. Similarly, CrackSAM, despite its superior semantic modeling, shows reduced ability to capture small-scale crack details in 3DCrack, which are less distinguishable from the background in a zoomed-out context, as shown in Tab.~\ref{tab:exp1} and Fig.~\ref{fig:exp1}.

\begin{figure}
    \centering
    \includegraphics[width=1\linewidth]{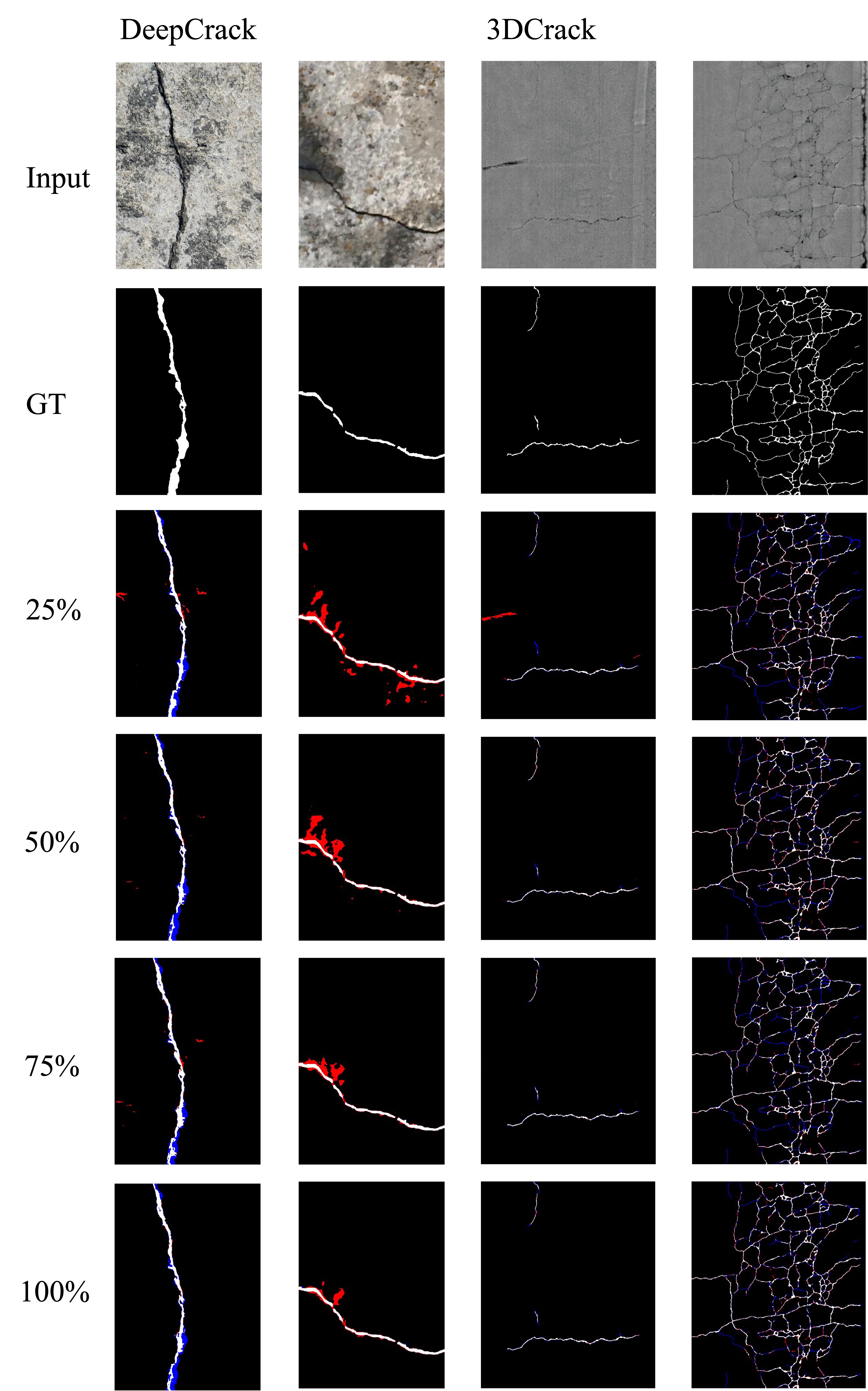}
    \caption{Crack segmentation results on DeepCrack and 3DCrack datasets from ResUNet with different training quantities. False positives are highlighted in red, and false negatives are highlighted in blue.}
    \label{fig:exp2}
\end{figure}

\textbf{Training Quantity Effect.} The impact of training data quantity is clearly visible in Tab.~\ref{tab:exp1} and the qualitative results in Fig.~\ref{fig:exp2}. All models benefit significantly from more labeled data, improving their ability to correctly segment cracks, suppress noise, and handle diverse crack patterns. In particular, an increase in training data helps models better differentiate true cracks from spurious patterns. In Fig.~\ref{fig:exp2}, this improvement is visually represented: the red regions (false positives) shrink, while the blue regions (false negatives) are increasingly replaced by white (true positives), indicating enhanced model accuracy and recall as data volume increases.

Experimental results also indicate that the performance of most models tends to plateau when the training data is increased from 75\% to 100\%, suggesting that exploring more efficient strategies for leveraging additional data remains an important direction for future research.

\textbf{Low-Data Performance.} CrackSAM and UniMatchv2 stand out in scenarios with limited labeled data. UniMatchv2 excels due to its semi-supervised learning framework and the use of the powerful DINOv2 backbone, which offers richer feature representations compared to the ResNet-based backbone in UniMatchv1. CrackSAM also performs well, benefiting from fine-tuning on the strong prior knowledge embedded in the SAM foundation model. Its pretrained capability in object-level semantic understanding makes it particularly effective in identifying cracks with minimal supervision. These results suggest that models incorporating pretrained semantic priors or designed for semi-supervised learning have a clear advantage in data-scarce conditions.

\subsection{Generalizable Experiments}
\label{sec:generalizable}

\begin{table}
\centering

\caption{Performance of Generalizable Models on different datasets trained with combined datasets. The \textbf{best performance} is highlighted in bold, and the \underline{second best} is underlined.}
\label{tab:exp2}

\label{tab:results}
\resizebox{\columnwidth}{!}{%
\begin{tabular}{llclll}
\toprule
 \multirow{2}{*}{Method}  & \multirow{2}{*}{Metric}  &\multicolumn{2}{c}{Zero Shot} & \multicolumn{2}{c}{Many Shot}\\ 
  \cmidrule(lr){3-4} \cmidrule(lr){5-6}
  &  & DeepCrack&3DCrack & DeepCrack&3DCrack \\ 

\midrule

 ResUNet &  \multirow{4}{*}{IoU}& 0.441&\underline{0.279}& \underline{0.633}&\textbf{0.67}\\
 DeepCrack &  & 0.387&0.181& 0.557&0.584\\

 CrackSAM &  & \underline{0.502}&\textbf{0.3}& \textbf{0.711}&\underline{0.6}\\ 
Cycle GAN & & \textbf{0.621}&-- & --&-- \\

\midrule

 ResUNet &  \multirow{4}{*}{Precision}& \underline{0.589}&\underline{0.298}& \textbf{0.848}&\textbf{0.824}\\
 DeepCrack &  & 0.64&0.207& 0.793&\underline{0.723}\\

 CrackSAM &  & 0.551&\textbf{0.319}& \underline{0.834}&0.699\\ 
 Cycle GAN & & \textbf{0.776}&-- & --&-- \\
 \midrule

  ResUNet &  \multirow{4}{*}{Recall}& 0.731&\underline{0.729}& \underline{0.736}&\underline{0.732}\\
 DeepCrack &  & 0.602&0.633& 0.681&0.682\\

 CrackSAM &  & \textbf{0.903}&\textbf{0.758}& \textbf{0.846}&\textbf{0.742}\\ 
 Cycle GAN & & \underline{0.768}&-- & --&-- \\
 \midrule

   ResUNet &  \multirow{4}{*}{Dice}& 0.568&\underline{0.375}& \underline{0.742}&\textbf{0.767}\\
 DeepCrack &  & 0.514&0.27& 0.674&0.686\\

 CrackSAM &  & \underline{0.644}&\textbf{0.392}& \textbf{0.816}&\underline{0.711}\\ 
 Cycle GAN & & \textbf{0.745}&-- & --&-- \\
 \bottomrule

\end{tabular}
}

\end{table}

\subsubsection{Experimental Settings}

In this experiment, ResUNet~\cite{ronneberger2015u,he2016deep} and DeepCrack~\cite{liu2019deepcrack} are trained the same way as in the previous experiment. CrackSAM~\cite{ge2024fine, kirillov2023segment} and CycleGAN~\cite{zhu2017unpaired} are trained following the original papers. There are two different settings, zero shot and many shot, in this generalizable experiments. The test set remains the same, composed of test sets from DeepCrack~\cite{liu2019deepcrack} and 3DCrack. The training set for zero-shot experiments does not include any images from either datasets, while the training sets from both datasets are included in the training data in many-shot setting.

\textbf{Zero Shot.} For the zero-shot study, we explore two distinct groups of approaches commonly adopted in the literature. The first group focuses on leveraging large-scale datasets to enable models to generalize across diverse scenarios through extensive data exposure~\cite{ge2024fine}. These approaches typically rely on high-capacity architectures with larger parameter counts, allowing them to learn complex feature representations from a wide variety of input conditions. In our experiments, we evaluate three such models, ResUNet~\cite{ronneberger2015u,he2016deep}, DeepCrack~\cite{liu2019deepcrack}, and CrackSAM ~\cite{ge2024fine, kirillov2023segment}, which represent different families of large-scale segmentation networks known for their strong generalization performance when trained on sufficiently diverse data from Khanh11k~\cite{Khanh} with data belonging to DeepCrack removed for zero-shot purpose. We also include CycleGAN~\cite{zhu2017unpaired} to evaluate how well the unpaired image-to-image translation can produce reliable segmentation maps from real-world inputs in an unsupervised manner; specifically, we shuffle the annotation to create unpaired training data based on curated Khanh11k mentioned above.

\begin{figure}
    \centering
    \includegraphics[width=1\linewidth]{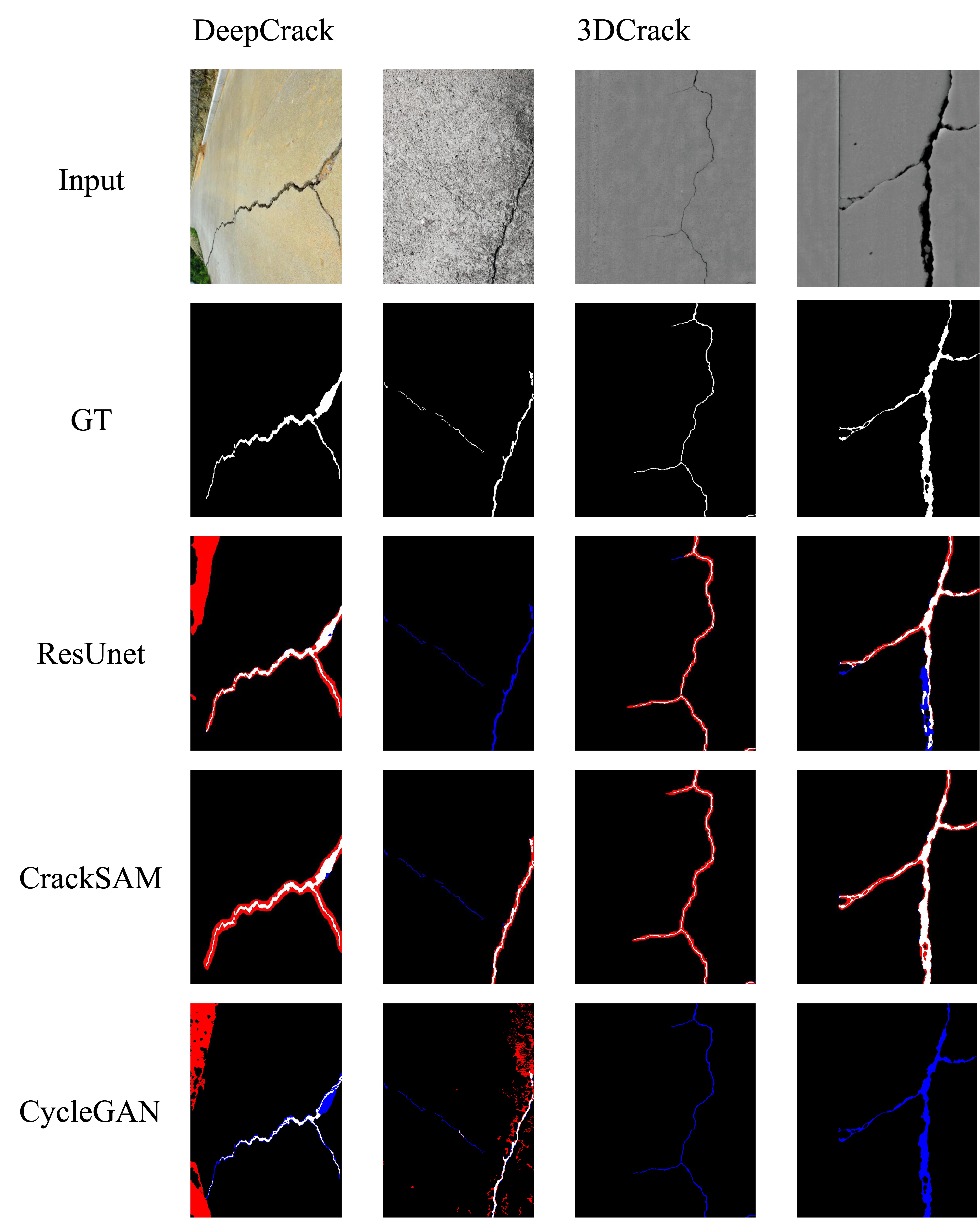}
    \caption{Zero-shot performance from ResUNet, CrackSAM, and CycleGAN. False positives are highlighted in red, and false negatives are highlighted in blue.}
    \label{fig:exp3}
\end{figure}

\textbf{Many Shot.} To further explore the impact of data diversity and volume on segmentation performance, we conduct a many-shot experiment by augmenting the training set of the Khanh11k dataset~\cite{Khanh} with additional labeled samples from the whole training subsets in both DeepCrack and 3DCrack datasets. This setup allows us to simulate a realistic, fully supervised training scenario where the model has access to rich, heterogeneous data sources. The objective is to assess whether exposure to more diverse crack appearances and imaging conditions can improve the generalization and robustness of crack segmentation models.

\begin{figure}
    \centering
    \includegraphics[width=1\linewidth]{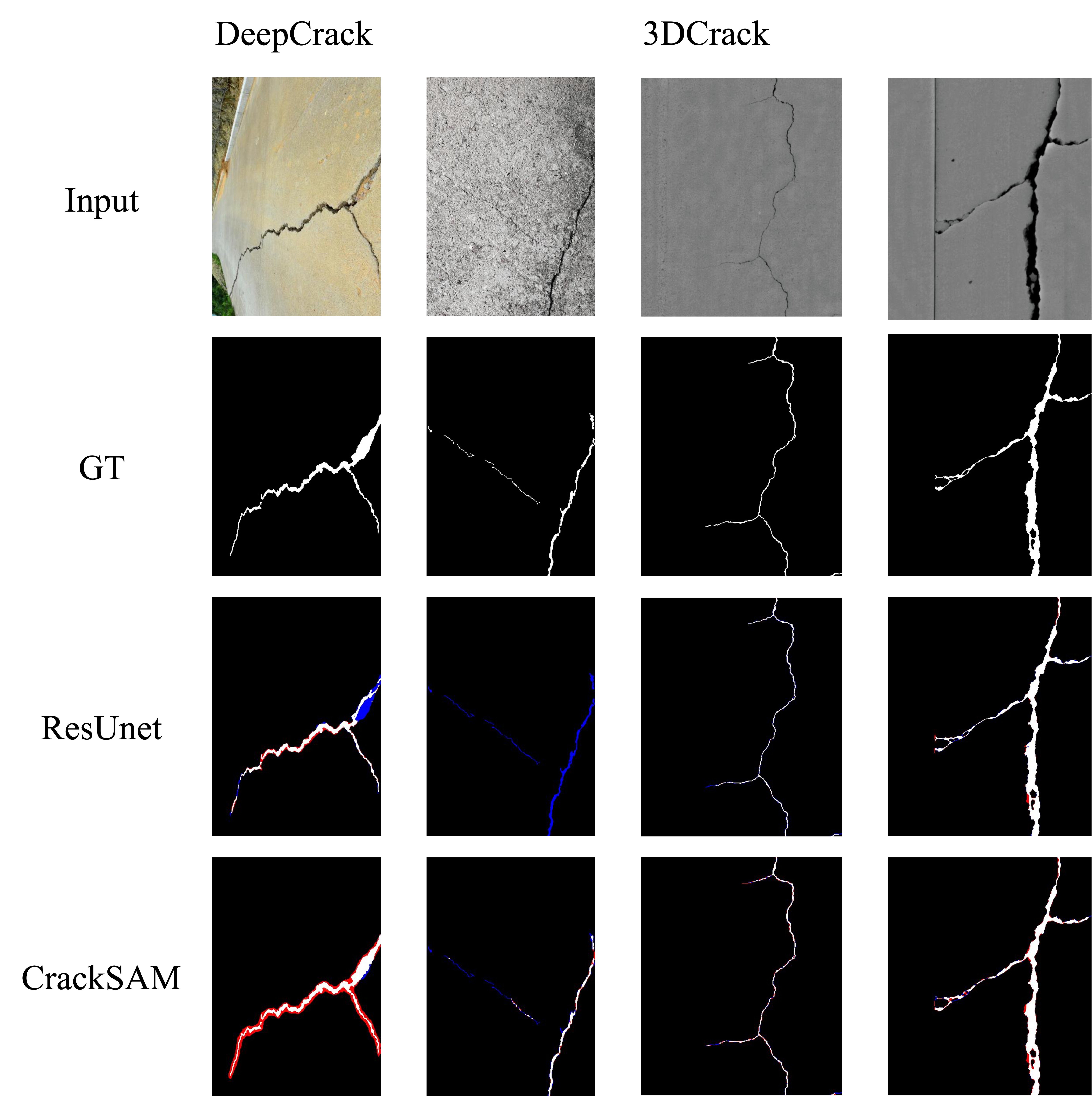}
    \caption{Many-shot performance from ResUNet and CrackSAM. False positives are highlighted in red, and false negatives are highlighted in blue.}
    \label{fig:exp4}
\end{figure}

\subsubsection{Findings}

\textbf{Zero Shot.} As shown in Tab.~\ref{tab:exp2} and Fig.~\ref{fig:exp3}, all selected models demonstrate the ability to perform zero-shot crack segmentation. However, despite some positive results, zero-shot performance remains limited due to generalization gaps, particularly in the presence of domain shifts. The most prevalent issue observed is the overshoot and undershoot phenomenon: models either miss fine crack structures (undershoot) or mistakenly segment non-crack regions due to visual similarity (overshoot).

ResUNet, which performs strongly in supervised scenarios, struggles in this setting. It frequently fails to identify the correct crack regions and is easily distracted by background patterns such as surface textures, road markings, or shadows. This suggests that without dataset-specific fine-tuning, its ability to generalize semantic boundaries is limited.

CrackSAM demonstrates more robust segmentation compared to ResUNet in this zero-shot context. Owing to its foundation in the SAM, it shows improved localization of crack areas and exhibits fewer distractions. However, it still exhibits a consistent tendency to over-segment regions surrounding the cracks, leading to inflated boundaries and occasional inclusion of irrelevant pavement details.

CycleGAN delivers a strong performance on the DeepCrack dataset among the three models in this setting. As an unsupervised image-to-image translation method, CycleGAN leverages domain adaptation to stylize input images to resemble the training distribution, thus enabling it to bridge visual gaps between domains. This helps it to more closely approximate crack patterns seen during training. However, CycleGAN’s approach is purely style-driven without a task-specific objective, so it remains vulnerable to noise and false positives, particularly in areas with texture similarities to cracks. Moreover, its performance collapses on the 3DCrack dataset, where cracks are thinner, fainter, and embedded in a large non-crack background. This indicates a critical weakness of style transfer-based models in dealing with fine-grained structural features that lack strong visual saliency.

In summary, while zero-shot segmentation offers an attractive direction for reducing annotation burdens, current methods still face significant challenges in accurately handling crack boundaries and adapting to new domains with different visual characteristics. Future work may focus on integrating prior knowledge~\cite{lin2023structtoken,huo2022domain}, uncertainty modeling~\cite{fleuret2021uncertainty,li2022uncertainty}, or contrastive learning strategies~\cite{radford2021learning,luddecke2022image} to enhance zero-shot robustness and mitigate overshoot/undershoot errors.

\textbf{Many Shot.} As illustrated in Tab.~\ref{tab:exp2} and Fig.~\ref{fig:exp4}, all models demonstrate substantial improvements in segmentation accuracy compared to the zero-shot setting. The performance gains are particularly evident in terms of Dice score and IoU, with fewer false positives in non-crack regions and better delineation of crack boundaries. 

Moreover, all models become more proficient at identifying finer cracks, which were previously under-segmented in zero-shot conditions. This indicates that the increased volume and variation of annotated cracks in the training data improves the model’s ability to capture subtle visual cues that define complex or less salient crack patterns.

However, despite these improvements, when comparing the results in Tab.~\ref{tab:exp2} (many-shot) with those in Tab.~\ref{tab:exp1} (individual dataset training with full supervision), there is no performance gain achieved. This suggests that simply increasing the quantity and diversity of training data from multiple domains does not automatically lead to proportional performance improvement. One possible reason is the domain discrepancy among datasets: DeepCrack, 3DCrack, and other datasets in Khanh11k differ in imaging perspectives, crack styles, background textures, and distraction types. Baseline models may still struggle to reconcile these differences, underscoring the need for task-specific designs, including tailored architectures, loss functions, etc.

These findings point to an important direction for future research: developing models that can more effectively utilize diverse and domain-varied training data. This may involve techniques such as feature alignment~\cite{chen2019progressive,yeh2021sofa,radford2021learning}, curriculum learning~\cite{zhang2017curriculum,soviany2022curriculum} and self-supervised learning~\cite{hoyer2023mic} that focus on invariant crack characteristics across datasets. Addressing this limitation will be key to building truly robust and generalizable crack segmentation systems.
\section{Open Challenges}
\label{sec:open}
Despite the rapid progress and promising results achieved in recent years, the task of automated crack detection remains far from being fully solved as shown in our review and experiments. In the following, we outline some of the key open challenges that continue to shape the research landscape in crack detection.

\subsection{Learning Paradigms}

\textbf{Efficient Utilization of Data.}
Our review has revealed a growing trend toward developing learning strategies that can make more efficient use of available data. Despite the ongoing progress in dataset expansion and the increasing adoption of large pre-trained models, there will always be edge cases, unseen scenarios, and variations that are not fully captured by existing data. Acquiring a comprehensive, high-quality dataset including new variations of a large quantity remains resource-intensive, especially when pixel-level annotations are required. To address these challenges, future research can further explore with data-efficient learning paradigms, such as SSL, WSL, USL, FSL, DA and foundation models with PEFT. These approaches have shown promising results and have the potential to further maximize model performance while minimizing the reliance on large labeled datasets, thereby enhancing scalability and fast adaptation in practical applications.

\textbf{Benchmarking and Standardization.}
As the field continues to explore a wide range of learning paradigms from supervised to foundation model adaptation, the need for standardized benchmarks becomes increasingly critical. Without consistent datasets, evaluation metrics, and task definitions, it is difficult to make fair comparisons between methods or draw reliable conclusions about their strengths and weaknesses. Currently, many studies use custom evaluation protocols, which hampers reproducibility and slows down collective progress. Establishing shared benchmark datasets with clearly defined training and test sets, standardized annotation formats, and unified performance metrics would enable more rigorous and transparent evaluations~\cite{tsai2012critical, tsai2017comprehensive}. This study contributes to this goal through the newly released 3DCrack dataset and tailored benchmarking designs. In addition, maintaining leaderboards and organizing challenges could further incentivize innovation and provide a centralized reference point for progress in crack detection and segmentation.

\subsection{Generalizability \& Model Development}

\textbf{Scaling to Larger Models and Foundation Architectures.}
Recent advances in computer vision have been driven by large-scale models and foundation architectures such as transformers~\cite{vaswani2017attention}, CLIP~\cite{radford2021learning}, DINO~\cite{seong2023leveraging}, and SAM~\cite{kirillov2023segment}, which have shown remarkable generalization across tasks and domains. Inspired by this trend, some preliminary efforts have been made to apply such foundation models to crack detection~\cite{ge2024fine}. Although the results are promising, this area is still underexplored and the overall performance is not perfect as we show in Sec.~\ref{sec:generalizable}. Specifically, false positives caused by distracting patterns and coarse detection should be eliminated to enable subsequent tasks such as crack quantization~\cite{hsieh2023pavement}.

Future work can investigate how to better adapt foundation models to crack detection across different scenarios, potentially by incorporating domain-specific fine-tuning~\cite{zhang2023customized,ge2024fine,hu2022lora,huang2025directional, yu2025fm}, prompt tuning~\cite{xiao2024cat}, or specialized modules for thin-structure segmentation~\cite{zhang2019net,Zhu_2024_CVPR}.

\textbf{Crack-Specific Architectural and Objective Designs.}
Unlike general objects, cracks are typically thin, elongated, and often discontinuous structures, making them challenging to segment using off-the-shelf models. Many studies have introduced crack-aware architectures or loss designs that take the geometric and topological features of cracks into consideration~\cite{zou2018deepcrack, zhang2018deep, hsieh2021dau, hsieh2023pavement}. These designs have demonstrated clear benefits in general performance. Similarly, continued research in this direction is necessary to develop models that explicitly account for the unique spatial patterns of cracking, decreasing false positives and false negatives for finer results.

\textbf{Real-Time and Efficient Architectures.}
As model sizes increase to achieve higher accuracy, there is a growing need to balance performance with computational efficiency, especially for deployment in real-world scenarios such as mobile inspection robots or devices. Real-time inference is crucial for practical applications. Lightweight backbone architectures (e.g. MobileNet~\cite{howard2017mobilenets}, EfficientFormer~\cite{li2022efficientformer}, CSegamba~\cite{liu2025scsegamba}), pruning~\cite{cheng2024survey}, quantization~\cite{polino2018model}, and knowledge distillation~\cite{polino2018model} are all viable directions to explore. Designing models that maintain competitive accuracy while achieving low latency and small memory footprints will be a key challenge moving forward.

\textbf{Multimodal Fusion.}
Crack detection has the chance to benefit from integrating data beyond standard RGB imagery. Multimodal fusion, for instance, can involve combining sensor data such as depth, infrared information~\cite{liu2022asphalt,zhou2023deep}. Fusing multiple modalities has the potential to improve robustness under varying environmental conditions and provide richer contextual understanding. Deep learning models capable of aligning and processing heterogeneous inputs, such as vision-language models~\cite{radford2021learning, zhang2024vision} or sensor/feature fusion networks~\cite{yeong2021sensor,fan2025robust, hsieh2023pavement}, represent a promising area for future research. Challenges include synchronization of data sources, modality-specific noise, and the need for large annotated multimodal datasets.

\subsection{Dataset}

\textbf{Data Scarcity.}
Despite the existence of several publicly available crack segmentation datasets, their size remains limited. These datasets typically contain a few hundred to a few thousand images, which is relatively small compared to large-scale datasets in other computer vision tasks (e.g., COCO~\cite{lin2014microsoft}, Cityscapes~\cite{cordts2016cityscapes}). To address this limitation, researchers have attempted to aggregate multiple datasets into a unified collection, yielding performance improvements by increasing training diversity and quantity. However, the overall volume and diversity of data still fall short compared with other domains, restricting the robustness and generalizability of models.

\textbf{Lack of Industry-Level Data.}
Another critical gap lies in the limited availability of professionally acquired and annotated datasets. Most public datasets rely on RGB imagery collected with consumer-grade cameras under uncontrolled conditions. In contrast, data acquired using specialized sensors (e.g., 3D laser scans) provides higher fidelity and is more representative of real-world inspection scenarios. Our work attempts to bridge this gap by releasing a dataset that features fine-grained, expert-annotated crack masks. Nevertheless, there remains a strong need for additional high-quality data from diverse environments and sensors. Considering the substantial effort required for pixel-level annotation, future research can also consider exploring the use of semi-automatic annotation tools~\cite{rother2004grabcut, kaul2011detecting, lin2016scribblesup} to scale up dataset generation.

\textbf{Data Quality and Annotation Methods.}
In addition to data quantity, quality is another concern. Manual annotations are inherently prone to subjectivity and inter-annotator variability. Factors such as different interpretations of what constitutes a crack, variations in perceived severity levels, and inconsistencies in annotation protocols can all contribute to noisy ground truth labels. Establishing clear annotation standards, evaluation metrics for quality assurance, and introducing consensus-based or probabilistic labeling strategies could help improve annotation consistency.

On the other hand, alternative annotation strategies deserve further exploration. Rough annotations offer a promising direction for accelerating data generation, especially when supported by learning paradigms such as weakly supervised learning. Incorporating experts in the loop may also be beneficial, as it not only ensures annotation quality but also facilitates continual learning through iterative feedback and model refinement.

\section{Conclusion}
\label{sec:conclusion}

In this paper, we present a comprehensive review of deep learning-based crack detection, focusing on seven major learning paradigms, the growing emphasis on model generalizability, and the diversification of datasets in the field. We categorize and discuss the shifts in learning paradigms with detailed subcategories and representative works, highlighting how the field has progressed from conventional supervised approaches to more data-efficient paradigms such as semi-supervised, few-shot methods, etc. The emergence of generalizability is identified as a key trend, reflecting the increasing need for models that can perform reliably across diverse and unseen conditions.

In terms of data, we observe a clear trend toward larger, more diverse, and sensor-rich datasets. As part of this effort, this paper releases a high-resolution, large-scale, industry-level crack segmentation dataset 3DCrack with diverse pavement conditions, contributing a valuable resource to the community for benchmarking and future development.

Although trends in crack detection are closely connected with those in the broader CV community, the task demands specialized designs and considerations. Drawing from both the review and our empirical studies, we also identify and discuss open challenges that remain unresolved. These challenges, spanning learning paradigms, generalizability and model development, as well as dataset, outline critical directions for future research. We hope this work provides a useful reference for researchers and practitioners in this field and encourages further efforts toward robust, scalable, and generalizable crack detection systems.

\section{Acknowledgment}
The authors would like to thank Prof.~Zsolt Kira, Dr.~Mi Zhou, Jacob Biros from Georgia Institute of Technology, as well as Dr.~Muhammad Zubair Irshad from Toyota Research Institute for helpful discussion and valuable feedback.
{
    \small
    \bibliographystyle{ieeenat_fullname}
    \bibliography{main}
}

\end{document}